\documentclass[runningheads]{llncs}

\usepackage[mobile]{eccv}

\usepackage{eccvabbrv}

\usepackage{graphicx}
\usepackage{wrapfig}
\usepackage{multirow}
\usepackage{multicol}
\usepackage[dvipsnames]{xcolor}
\usepackage{booktabs}
\usepackage{bbding}
\usepackage{colortbl}

\usepackage[accsupp]{axessibility}  

\usepackage[pagebackref,breaklinks,colorlinks,citecolor=eccvblue]{hyperref}

\usepackage{orcidlink}

\begin{document}

\title{RaFE: Generative Radiance Fields Restoration} 


\author{Zhongkai Wu\inst{1} \and
Ziyu Wan\inst{2} \and
Jing Zhang\inst{1} \and
Jing Liao\inst{2} \and
Dong Xu\inst{3}}

\authorrunning{Wu et al.}

\institute{College of Software, Beihang University, China \and
City University of Hong Kong, China \and
The University of Hong Kong, China\\
\email{ZhongkaiWu@buaa.edu.cn} \\
\href{https://zkaiwu.github.io/RaFE}{zkaiwu.github.io/RaFE}
}

\maketitle

\begin{figure}[t]
  \centering
   \includegraphics[width=1.0\linewidth]{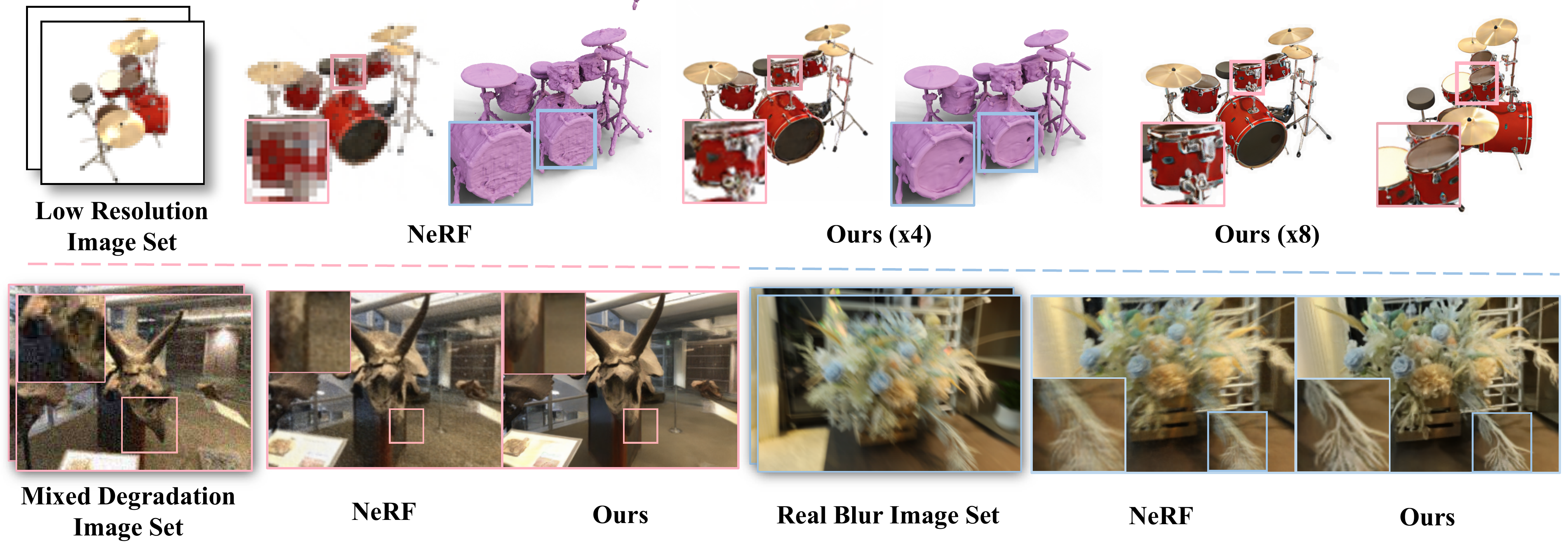}
   \vspace{-2em}
    \caption{\textbf{High-quality restoration of radiance fields from various types of degradation.} Given only degraded images, our method can restore high-quality NeRF. It is a generic approach that can be applied to various types of degradations, resulting in refinement of both geometry and appearance.}
   \label{fig:teaser}
   \vspace{-2.em}
\end{figure}

\begin{abstract}
NeRF (Neural Radiance Fields) has demonstrated tremendous potential in novel view synthesis and 3D reconstruction, but its performance is sensitive to input image quality, which struggles to achieve high-fidelity rendering when provided with low-quality sparse input viewpoints. Previous methods for NeRF restoration are tailored for specific degradation type, ignoring the generality of restoration. To overcome this limitation, we propose a generic radiance fields restoration pipeline, named RaFE, which applies to various types of degradations, such as low resolution, blurriness, noise, compression artifacts, or their combinations. Our approach leverages the success of off-the-shelf 2D restoration methods to recover the multi-view images individually. Instead of reconstructing a blurred NeRF by averaging inconsistencies, we introduce a novel approach using Generative Adversarial Networks (GANs) for NeRF generation to better accommodate the geometric and appearance inconsistencies present in the multi-view images. Specifically, we adopt a two-level tri-plane architecture, where the coarse level remains fixed to represent the low-quality NeRF, and a fine-level residual tri-plane to be added to the coarse level is modeled as a distribution with GAN to capture potential variations in restoration. We validate RaFE on both synthetic and real cases for various restoration tasks, demonstrating superior performance in both quantitative and qualitative evaluations, surpassing other 3D restoration methods specific to single task. Please see our project website \href{https://zkaiwu.github.io/RaFE/}{zkaiwu.github.io/RaFE}.
\vspace{-0.5em}
\end{abstract}

\vspace{-2em}
\section{Introduction}
\vspace{-0.5em}
\label{sec:intro}

Recently, Neural Radiance Fields (NeRFs)~\cite{mildenhall2021nerf, mueller2022instant, barron2021mipnerf, barron2022mipnerf360, Chen2022ECCV, fridovich2023k, SunSC22, Wan_2023_CVPR} have achieved great success in novel view synthesis and 3D reconstruction. However, most NeRF methods are designed based on well-captured images from multiple views with calibrated camera parameters. In real-world applications of NeRF, the data capture or transmission process often introduces various forms of image degradations, such as noise generated during photography in low-light conditions ~\cite{mildenhall2022rawnerf, pearl2022noiseaware} and blur caused by camera motion ~\cite{ma2022deblur, wang2023badnerf}, or JPEG compression and down-sampling during transmission~\cite{wang2022nerf, bahat2022neural}. Simply restoring degraded images frame-by-frame can result in inconsistencies of geometry and appearance across different viewpoints. Directly reconstructing 3D models over these per-frame restoration results can easily induce inferior quality since current NeRF methods heavily rely on pixel-wise independent ray optimization with local computations, which are highly vulnerable to noise and other degradation.

Several NeRF variants have attempted to reconstruct 3D scenes with degraded multi-view images by introducing specific strategies or additional constraints for the optimization of radiance fields. For example, \cite{ma2022deblur, lee2023dp, lee2023exblurf} deal with image blur artifacts by modeling the degradation kernel with NeRF, while \cite{wang2022nerf, bahat2022neural} super-sampling on rays or tri-planes to obtain high-resolution 3D from low-resolution observations. 
Additionally, \cite{mildenhall2021nerf} modifies NeRF to reconstruct the scene in linear HDR space by supervising directly
on noisy raw images to address the noise generated in low-light conditions.
\cite{zhou2023nerflix} tries to improve the view synthesis quality by removing NeRF-specific rendering artifacts. Even with great success, all these approaches are only designed to handle specific types of degradation. To the best of our knowledge, currently there is no \textit{generic pipeline} which supports the restoration of radiance fields under various types of degradation. 

In this paper, we propose RaFE, a generic NeRF restoration framework that enables high-quality radiance fields reconstruction from captured images containing various types of degradation in a generative manner. Firstly, we leverage the success of off-the-shelf image restoration methods to restore the multi-view images with different forms of degradation, such as super-resolution, deblurring, denoising, removing compression artifacts or a combination thereof. It is important to note that since the images from different views are independently restored, there inevitably exist geometric and appearance inconsistencies between them. Naively optimizing a radiance field with these refined images would average out the inconsistencies and result in blurry outputs. To overcome this challenge, our insight here is, instead of describing a single 3D using inconsistent frames, we could consider these restored multi-view images as the renderings from multiple distinct high-quality NeRF models with varied geometry and appearance. In this case, we abandon the commonly-used pixel-wise reconstruction objective and propose to leverage the generative adversarial networks (GANs) to model the distribution of these different high-quality NeRF models, which could effectively capture the inherent variability in ill-posed inverse problem, allowing for a better accommodation of the inconsistencies present in different views.

Specifically, our pipeline consists of two main stages.  In the first stage, based on the type of degradation, we can employ the corresponding off-the-shelf image restoration methods~\cite{lin2023diffbir, saharia2022photorealistic, zhang2021designing, wang2022zero, saharia2022palette, kawar2022denoising} to obtain a set of high-quality multi-view images. In practice, we prefer choosing restoration methods
which have strong capabilities to recover high-quality and realistic texture details. 
In the second stage, we train a 3D generative model based on these restored multi-view images. Drawing inspirations from recent 3D generation works~\cite{chan2022efficient, bib:mimic3d, skorokhodov2022epigraf, wan2023cad}, we construct a convolutional neural network (CNN) to generate tri-plane features, which are subsequently sampled and decoded into density and colors using MLP networks for NeRF rendering. Here, instead of generating single-level tri-plane features as previous works did, we decompose the tri-planes into two levels. The coarse-level tri-planes are constructed directly from low-quality images and remain fixed during training, representing the coarse structure of the modeled 3D distribution. Simultaneously, we train a generator to output the diverse fine-level tri-plane features, which act as residuals to be added to the coarse-level features for NeRF rendering. By focusing on learning the residual representations instead of the entire tri-planes for NeRF, we simplify the modeling and learning of restoration variations 
since we only need to learn the details while the coarse structure is provided by coarse-level tri-planes, which makes great improvement in rendering quality for more complex regions.
To train the generator, we adopt an adversarial loss defined on NeRF rendered 2D images to encourage them to be indistinguishable from the high-quality restored images. We also incorporate a perceptual loss between the rendered images and the restored images to calculate structure constraints. 
Additionally, we propose patch sampling strategies to stabilize the generator training procedure. 
Once the generator has been trained, we can generate restored radiance fields with high quality renderings and a certain level of diversity by sampling different code in the latent space.


We conducted extensive experiments to validate the effectiveness of our method, both qualitatively and quantitatively. The experimental results showcase the superiority of our approach in various restoration tasks, such as superresolution (upper row of Figure~\ref{fig:teaser}), camera motion blur (a real-world case at the right part of the lower row of Figure~\ref{fig:teaser}) and the restoration of mixed degradation consisting of noise, blur, and compression (left part of lower row of Figure~\ref{fig:teaser}).  
Our method not only generates images with richer and enhanced texture details but also achieves significant improvements in geometric refinement, as demonstrated by the mesh visualization in Figure~\ref{fig:teaser}. To summarize, our contributions are:

\begin{itemize}
    \item We propose a generic radiance fields restoration pipeline that is applicable to various types of degradation.
    \item We introduce a generative method for NeRF restoration that enables better accommodation of geometric and appearance inconsistencies present in the multi-view images, thus allowing us to incorporate the success of image restoration into 3D restoration.
    \item  We show the restoration method performs well on various degradation scenarios with both enhanced appearance and refined geometry.
\end{itemize}

\section{Related Works}

\subsection{2D image restoration}
\vspace{-0.5em}

Image restoration is a long standing problem in low-level vision domain and significant progress has been achieved in different specific tasks including image super-resolution, deblur, denoise and blind restoration. Previously reconstruction-based methods ~\cite{zhou2023srformer, li2023real, chen2023dual, Li_2023_CVPR, zhang2023seal} show their success in these tasks. However, those reconstruction-based method are struggling to generate abundant high-quality details.
Subsequently, generative restoration methods\cite{lin2023diffbir, saharia2022photorealistic, zhang2021designing, wang2022zero,wan2020bringing, saharia2022palette, kawar2022denoising, chen2023hierarchical, tian2024visual}, particularly those based on diffusion model, have shown the great capability to render high-quality details. Deepfloyd \cite{saharia2022photorealistic} proposes a super-resolution model, which concatenates the low-resolution input with random noise at pixel level as a condition to guide the generation of high-resolution images. For blind restoration, DiffBIR~\cite{lin2023diffbir} designs a degraded pipeline to simulate real-world degradation and utilizes the pre-trained diffusion model to generate photorealistic images. For camera motion blur, HiDiff~\cite{chen2023hierarchical} recovers exquisite images by using diffusion to generate feature with abundant detailed information.

\vspace{-0.5em}
\subsection{Radiance Fields Restoration}
\vspace{-0.5em}

 NeRF restoration aims to reconstruct high-quality NeRF given only degraded images with various artifacts such as blur, noise, or low resolution. Up to now, several works~\cite{ma2022deblur,wang2023badnerf,lee2023exblurf,lee2023dp,pearl2022noiseaware,chen2023dehazenerf,mildenhall2021nerf,wang2022nerf,bahat2022neural,zhou2023nerflix,han2023super}
 have explored this task under specific type of degradation. \cite{ma2022deblur, wang2023badnerf, lee2023exblurf} deal with blurred input images by designing blur kernel or optimizing camera paths for NeRF rendering process. 
 For NeRF super-resolution, \cite{wang2022nerf} increased the ray sampling density, forcing multiple rays to render pixels equal to the same pixel, and applied a 2D refinement model to get final output images. \cite{han2023super} introduces a CEM~\cite{bahat2020explorable} refinement model to adjust the output of off-line super-resolution model for better multi-view consistency. However, the CEM refinement model ruins the structure details of the images and inconsistencies still exist, leading to a smooth reconstruction result. \cite{pearl2022noiseaware, mildenhall2021nerf} mainly focus on the noise degradation of the input image. \cite{mildenhall2021nerf} modifies NeRF to reconstruct the scene in linear HDR color space by supervising directly
on noisy raw input images to address the noise generated during post-processing from HDR images to LDR images, while \cite{pearl2022noiseaware} achieve NeRF restoration on noisy images by using noise-awarded encode to aggregate features across views. \cite{zhou2023nerflix} considers solving the degradation that occurs in NeRF reconstruction by training a 2D refinement model using simulated degraded images for typical NeRF-style artifacts, but they can not achieve refinement on geometry since the restoration only happens on rendered views. None of the existing approaches can restore NeRF in 3D space directly with flexible forms of degradation. NVSR~\cite{bahat2022neural} archives 3D geometry refine by upsampling tri-plane representation, but their training processing requires tremendous amounts of 3D data, which are extremely hard to obtain in practice. By contrast, our method has the ability to handle more flexible forms of degradation and restore 3D geometry and appearance with the only needs of an image set for an object or scene, making the 3D restoration more practical in real-world applications.

\section{Method}
\vspace{-0.25em}

\begin{figure}[t]
  \centering
   \includegraphics[width=1.0\linewidth]{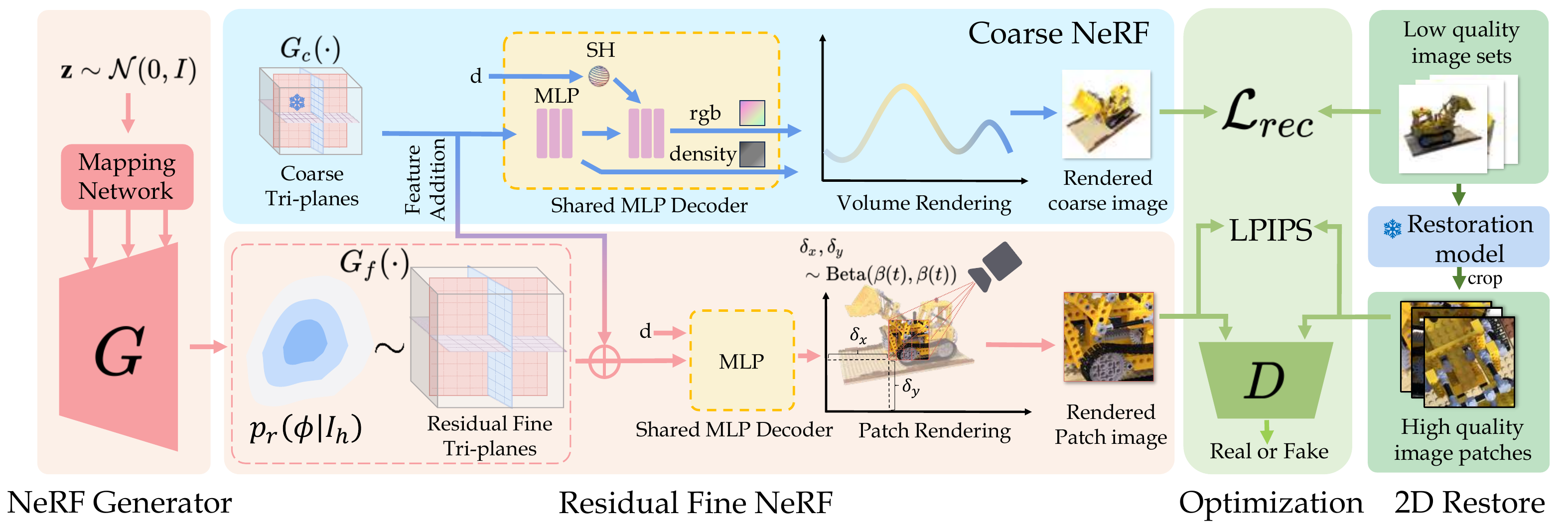}
   \vspace{-2em}
    \caption{\textbf{Overview of our pipeline.} Given multi-view degraded images, we utilize the off-the-shelf methods to restore high-quality multi-view images. Then, we train our Generative Restoration NeRF to generate a high-quality scene.}
    \vspace{-1.5em}
   \label{fig:method}
\end{figure}

In this section, we will elaborate the details of RaFE. We introduce how to refine the degraded views using pretrained 2D restoration model, to capture the high-quality appearance distribution in Sec.~\ref{2d_restoration}. Then, we describe our generative restoration framework including the neural representation, generator architecture and optimization in Sec.~\ref{3d_restoration}. The training strategy is introduced in Sec.~\ref{patch_strategy}. The overall pipeline could be found in Fig.~\ref{fig:method}.

\vspace{-0.5em}
\subsection{High-quality Image Restoration}\label{2d_restoration}

The recent success on 2D image restoration task is dominated by the denoising diffusion probabilistic model, benefiting from its powerful appearance generation capability and prior. Hence in this paper we mainly borrow the recent diffusion-based restoration methods~\cite{saharia2022photorealistic, lin2023diffbir} to obtain the high-quality multi-view images from their low-quality counterparts. It should be noted that our 3D restoration framework could also work pretty well based on non-diffusion restoration approaches. Please check the experimental parts for more details.  

To assist in the restoration of image details, we employ powerful image caption models~\cite{li2022blip, big_vision} to get an accurate textual prompt for the scene, denoted as $P$. Given a set of multi-view low-quality images $I_l$, the textual prompt is produced by selecting a view that contains as much information of the scene as possible, e.g. a side view of a synthetic Lego model. The prompt $P$ and low-quality images $I_l$ are then fed into the restoration model~\cite{saharia2022photorealistic, lin2023diffbir} to achieve high-quality per-frame refinement. We denote the restored high-quality images as $I_h$.

The diffusion model-based image restoration methods can effectively reconstruct high-quality images from various real-world degradation types. Moreover, thanks to the powerful generalization ability of diffusion models preserved in these restoration methods, we can handle the scene restoration in open domain without the needs of re-training. 
However, though the restored images have better quality, more variations among different views occur due to the generative nature of these restoration models, 
leading to serious multi-view inconsistencies.
Next, we focus on dealing with this issue with the proposed generative pipeline.

\vspace{-0.5em}
\subsection{Generative NeRF Restoration}\label{3d_restoration}
To deal with geometric and appearance inconsistencies across views, we treat these restored multi-view images distributed similarly to the rendered images from multiple slightly different high-quality NeRF models. Consequently, instead of directly fitting a NeRF model using the refined high-quality views which usually leads to blurry reconstruction, we are trying to learn the distribution of these diverse NeRF models by leveraging a generative method, allowing us to sample distinct restored 3D under the same degraded inputs. 

\noindent\textbf{3D Representation.} Following the recent 3D generative model \cite{chan2022efficient, bib:mimic3d, skorokhodov2022epigraf}, we adopt the hybrid explicit-implicit tri-plane representation for the feature field. This representation combines both explicit and implicit components to effectively model the density and RGB values. More specifically, to obtain the descriptor for any query location $\boldsymbol{x} \in \mathbb{R}^3$, we project the point onto three planes $(\boldsymbol{P}_{xy}, \boldsymbol{P}_{yz}, \boldsymbol{P}_{zx})$ to retrieve the corresponding features $(\boldsymbol{f}_{xy}, \boldsymbol{f}_{yz}, \boldsymbol{f}_{zx})$ using interpolation. Then we calculate the mean value of these three sampled features to obtain the final feature representation.

Once we obtain the features for a point along the ray, we use two MLPs to decode the density and RGB values. The first MLP, denoted as $\mathcal{M}_{dens}$, maps the features to the density value $\sigma \in \mathbb{R}$ and a color feature $\boldsymbol{f}_{color}$. The second MLP, denoted as $\mathcal{M}_{color}$, takes the color feature and the view direction as inputs and maps them to the RGB value $\boldsymbol{c} \in \mathbb{R}^3$. We empirically show that incorporating the view direction allows us to effectively model view-dependent effects, particularly when there are non-Lambertian surfaces in the scene. After obtaining the densities and RGB values for each sampled point along the emitted ray, we can apply the classic volumetric rendering to obtain the color value for each pixel: 
\vspace{-0.5em}
\begin{equation}
 \vspace{-0.5em}
 \small
    \begin{split}
    C(\boldsymbol{r}) = \sum^{N}_{i=1}T_i(1-exp(-\sigma _i\delta _i))\boldsymbol{c}_{i}, 
     \quad
    T_i = exp(-\sum ^ {i-1}_{j=1}\sigma _i\delta _i),
  \label{eq:render}
  \end{split}
\end{equation}
where $\boldsymbol{c}_i, \sigma_{i}$ represents RGB and density value of the $i_{th}$ point alone the ray, respectively. $N$ is the sampled number and $\delta _i$ is the distance between samples.


\noindent\textbf{NeRF Generator.}
Given restored multi-view images, we regard them as renderings from several diverse high-quality NeRF models with varied geometry and appearance. The distribution formed by these distinct NeRF models, formulated as $p_r( \mathbf{\phi} |I_h)$, can be modeled by a generator.
With the hybrid explicit-implicit tri-plane representation, we introduce a StyleGAN2~\cite{Karras2021}-like CNN-based generator, which receives a latent code $\boldsymbol{w}$ mapped from random code $\boldsymbol{z}$ to generate fine tri-planes $\boldsymbol{P}_{f}$ for high-quality NeRF rendering. To fully leverage the degraded images and make the generator only focus on generating the necessary refinements, we also pre-train a coarse NeRF model with tri-plane features denoted as $P_{c}$ using the input low-quality images. Specifically, the coarse NeRF model consists of a coarse tri-plane $\boldsymbol{P}_{c}$ features and a decoder (where the decoder of coarse tri-planes is shared with the one of fine tri-planes). To obtain the final refined tri-planes $P$, we combine the reconstructed tri-plane representations with the residual tri-planes generated by the generator via: $\boldsymbol{f} = \boldsymbol{f}_c + \boldsymbol{f}_f$. By merging these components, RaFE effectively captures both the global geometric guidance provided by the coarse NeRF and the local refinements learned from 3D generator, enabling us to obtain the restored tri-planes that exhibit enhanced geometric accuracy and appearance fidelity.

\noindent\textbf{Optimization.}
To supervise the generator and NeRF parameters, we propose to minimize the distribution discrepancy between the rendered images and the restored high-quality images.
We adopt a saturate GAN~\cite{goodfellow2014generative} loss with an image level discriminator. Specifically, we treat the high-quality images restored by 2D model as the real samples while the rendered images as the fake samples, and utilize adversarial loss with R1 regularization between the real and fake images:
\vspace{-1.5em}
\begin{equation}
    \vspace{-0.5em}
    \begin{split}
    \mathcal{L}_{adv} &= \mathbb{E}_{\boldsymbol{z} \sim p_z, \boldsymbol{\theta} \sim p_{\theta}}[f(D(G(\boldsymbol{z, \boldsymbol{\theta}}))] \\ 
     & + \mathbb{E}_{I_{h} \sim p_{h}}[f(-D(I_{h})) + \lambda \| \nabla D(I_{h})  \|^2],
  \label{eq:l_mimic}
  \end{split}
\end{equation}
where $\boldsymbol{z}, \boldsymbol{\theta}$ represent random code and view point, respectively, and $I_h$ is the restored high quality image.


However, we observed relying solely on a GAN loss for training can lead to significant geometric mismatches between the restored images and rendered views. We argue that although the GAN loss helps align the distribution of 2D renderings, it still lacks geometry-level constraints. Therefore, we also incorporate a perceptual loss that encourages the rendered images to resemble the geometry of pre-frame restoration.
\vspace{-0.5em}
\begin{equation}
  \vspace{-0.5em}
  \mathcal{L}_{geometry} = LPIPS(I_{h}, G( \boldsymbol{z}, \boldsymbol{\theta})),
  \label{eq:mimic_loss}
\end{equation}
where $LPIPS(\cdot, \cdot)$ refer to the learned perceptual image patch similarity proposed in \cite{zhang2018perceptual}.
$I_{h}$ is the restored high quality image paired with view point $\theta$, and $G( \boldsymbol{z}, \boldsymbol{\theta})$ are the rendered image using the same training view $\boldsymbol{\theta}$ and a random sampled latent code $\boldsymbol{z}$. We also supervise the coarse NeRF branch by the input RGB degraded images:
\vspace{-1.0em}
\begin{equation}
    \vspace{-0.75em}
    \begin{split}
    \mathcal{L}_{rec} = \mathbb{E}_{\boldsymbol{\theta} \sim p_{\theta}} [\|  G_{c}(\boldsymbol{\theta})-I_{l}^{\theta}\|^2],
  \label{eq:l_coarse}
  \end{split}
\end{equation}
where $p_{\theta}$ indicate the view point distribution, and $I_l^{\theta}$ is the low quality image corresponding to view point $\boldsymbol{\theta}$. Overall, the complete training objective is:
\vspace{-0.5em}
\begin{equation}
    \vspace{-0.5em}
    \mathcal{L} = \lambda_{geometry}\mathcal{L}_{geometry} + \lambda_{adv}\mathcal{L}_{adv} + \lambda_{rec}\mathcal{L}_{rec},
  \label{eq:l_tot}
\end{equation}
where $\lambda_{geometry}, \lambda_{adv}, \lambda_{rec}$ are trade-off parameters.

\vspace{-0.25em}
\subsection{Patch-based Training Strategy}\label{patch_strategy}
\vspace{-0.25em}
During the training process, fully rendering the entire image, such as $256^2$ or $512^2$ pixels, and then feeding it into the discriminator can be computationally intensive and resource-consuming. This is because volume rendering requires the computation of density and color values for sampling points along the ray for each pixel, which can become prohibitively expensive for a whole image.
Therefore, we only render a patch of the view at a time (i.e. $64^2$).
The rendered patches and high-quality image patches are randomly selected, and the discriminator in Eq.~\ref{eq:l_mimic} receives patches of rendered images and patches of high-quality images as fake images and real images respectively. For the perceptual loss in Eq.~\ref{eq:mimic_loss}, the rendered patches and the high-quality patches have the same spacial coordinate on their original images to guarantee that the cropped patches have the same semantic and local structure. 

One limitation of using patches instead of entire images for training is that it can be challenging to sample the local scenes evenly at the boundary regions of images, particularly for forward-facing data. The imbalanced training distribution will result in mode collapse during the training process.
To mitigate this issue, we employ a beta sampling strategy to determine the positions of the sampled patches. This strategy ensures that patches in the boundary regions of the image are adequately sampled.
More specifically, the beta sampling can be formulated as:
\vspace{-0.35em}
\begin{equation}
    \vspace{-0.25em}
    \delta_{x}, \delta_{y} \sim Beta(\beta(t), \beta(t)),
  \label{eq:beta_sample}
\end{equation}
where $\delta_{x}$ and $\delta_{y}$ denote the position offset in $x$ and $y$ directions respectively, while $\beta(t)$ are linearly annealed from $\beta(0) = 1$ to some final value $\beta(T)$ smaller than $1$. 
By using the beta sampling strategy, we can maintain a more balanced distribution of training data that focuses more on the boundary patches, alleviating mode collapse issue and improving the overall training stability.


\vspace{-0.25em}
\section{Experiments}

\vspace{-0.25em}
\subsection{Setup}
\vspace{-0.25em}

\noindent\textbf{Datasets.} We evaluate our model on the NeRF-Synthetic benchmark dataset~\cite{mildenhall2021nerf}, which contains 8 synthetic objects with images taken from different viewpoints uniformly distributed in the hemisphere. Following original setting, We hold 200 viewpoints for generating high-quality training data and 200 viewpoints for testing. Further, to demonstrate the generalization ability of our method, we also evaluate our method on complex real-world LLFF scenes~\cite{mildenhall2019llff} which consists of 8 scenes captured with roughly forward-facing images. We also demonstrate the superior performance of RaFE on real-world blur~\cite{ma2022deblur} and noise~\cite{mildenhall2022rawnerf} data.

\noindent\textbf{Evaluation Metrics.} Following the common practice of 3D reconstruction, we firstly try to evaluate each method with two standard image quality metrics: peak signal-to-noise ratio (PSNR), structural similarity index (SSIM)~\cite{wang2004image}, which however could not reflect the real 3D restoration performance according to our observations. Due to the generative characteristic of RaFE, the recovered radiance fields from RaFE is very high-quality, but may not faithfully follow the "ground-truth" 3D, since inversing a degraded signal is a highly ill-posed problem. Moreover, we also found the baselines with better scores over PSNR and SSIM still pose the degraded appearance with smooth texture details, as shown in Figure \ref{fig:psnr_and_others_compare}. Hence, a better way is to leveraging perceptual metric: learned perceptual image patch similarity (LPIPS)~\cite{zhang2018perceptual} which computes the mean squared error (MSE) between normalized features from all layers of a pre-trained VGG~\cite{simonyan2014very} encoder and is deemed to better correlate with human perception. Besides we also leverage the latest non-reference based image quality assessment metrics including LIQE~\cite{zhang2023liqe} and MANIQA~\cite{yang2022maniqa} to demonstrate the superior rendering quality of our method. 

\noindent\textbf{Implementation Details.}
We implement all the experiments by PyTorch. For the 2D generator and discriminator, we use a convolutional-based generator and discriminator used in StyleGAN2~\cite{Karras2019stylegan2}. In all experiments, we choose Adam optimizer for all the modules in our pipeline, with hyperparameters $\beta_1 = 0, \beta_2 = 0.99$. We use learning rate $2\times10^{-3}$ for both generator and discriminator. For loss weights, we use $\lambda_{mimic} = 0.5$, $\lambda_{adv}=1.0$, and $\lambda_{rec}=1.0$ for almost all experiments. We evaluate RaFE framework on 4 different 3D restoration tasks:
\begin{itemize}

  \item 4$\times$ Super-Resolution: On the blender dataset, we resize the image to $64\times64$ resolutions to get low-resolution images. As for LLFF data, we first center crop the training image to $188\times252$ to adapt to our $4\times$ super-resolution task and then resize to $47\times63$ to get low-resolution images. And we use Deepfloyd~\cite{saharia2022photorealistic} for 2D super-resolution.
  
  \item Deblur: On the LLFF dataset, we construct camera motion blur by applying the blur kernel following equation $I_{blur} = \boldsymbol{A} \ast I_{clear}, 
    \label{eq:noise_format}$ where $\boldsymbol{A}$ is the blur kernel and $\ast$ stands for the convolution operator. Different from the blur dataset proposed in Deblur-NeRF~\cite{ma2022deblur}, which contains blurred images with varying degrees of blur, and even includes a certain amount of high-resolution images, we apply a large blur kernel size (e.g. 13) and a more complex camera motion path to all the images in the dataset. We additionally construct consistent blur datasets by using the same blur kernel to the training image in a scene, which means the camera motion trajectory is the same for the training image set. The resolution of images is the same as the super-resolution task. We use HiDiff~\cite{chen2023hierarchical} to recover high-quality images.
  
  \item Denoise: On the LLFF dataset, we follow the noise model used in ~\cite{pearl2022noiseaware, mildenhall2018burst} and we get the noisy version of a clean image $I$ according to equation $I_{noisy}(x) = \mathcal N(I(x), \delta_r^2+\delta_s^2I^2(x)),$
where $\sigma_r$ is the signal-independent
read-noise parameter, $\sigma_s$ is the signal-dependent shot-noise. Following ~\cite{pearl2022noiseaware, mildenhall2018burst}, we use Gain level to represent the noise strength. We use gain levels $=8$ to get our noisy image. We use DiffBIR~\cite{lin2023diffbir} to get high-quality images.
  
  \item Mixed degradation: The degradation pipeline consists of three stages: \textbf{blur}, \textbf{noise}, and \textbf{JPEG compression}. First, we utilize Gaussian blur with a radius of 7 for the blender dataset and 3 for the LLFF dataset. Second, we add noise with std 25. And then, we apply JPEG compression. The quality of JPEG compression is 50. The resolution of images is the same as the super-resolution task. We use DiffBIR~\cite{lin2023diffbir} to get high-quality images.
\end{itemize}

\subsection{Results}

\noindent\textbf{Baseline Methods.}
For general tasks, we try a baseline that firstly restores the degraded images and uses the restored high-quality images to reconstruct a NeRF directly, denoted as NeRF-Perframe. We also use the 2D-based restoration model SwinIR~\cite{liang2021swinir} to do the per-view refinement for the renderings of NeRF trained by degraded image, denoted as NeRF-SwinIR. Note that we do not evaluate NeRF-SwinIR for the deblur task since there are no corresponding checkpoints.

To more thoroughly test the effectiveness of our method, we also select some task-specific competitors.
For super-resolution task, we choose NeRF-SR~\cite{wang2022nerf}, Neural Volume Super-Resolution (NVSR)~\cite{bahat2020explorable} as baselines. For mixed degradation, since there is no existing method tailored for mixed degradation, we choose NeRFLiX, which tries to solve the NeRF-like degradation by training a 2D refinement model using degradation images constructed by a degradation simulator for typical NeRF-style artifacts. We consider this to be the most relevant method. 
 For the deblur task, we compare with two state-of-the-art methods Deblur-NeRF~\cite{ma2022deblur} and BAD-NeRF~\cite{wang2023bad}. Deblur-NeRF designs a learnable blur kernel and applys it to rays to simulate the degrading process and BAD-NeRF directly models the camera trajectories to solve motion blur.
 For the denoising task, we compare with NAN~\cite{pearl2022noiseaware}, which uses a noise-aware encoder to aggregate the feature of multi-view images for restoration.

\begin{figure}[!t]
     \centering
     \begin{subfigure}[b]{0.43\textwidth}
         \centering
         \includegraphics[width=1.0\linewidth]{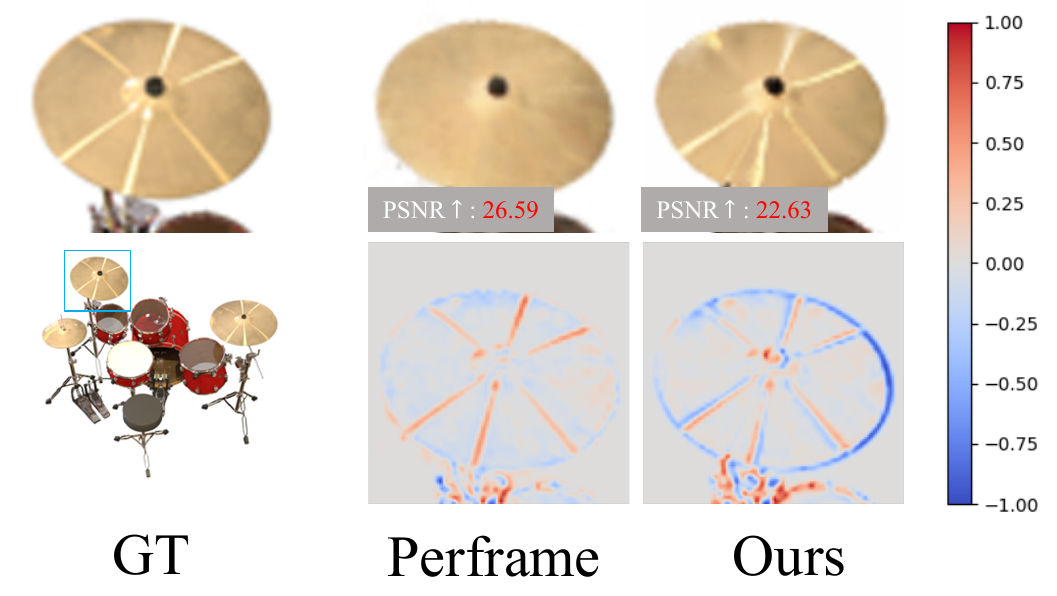}
         \caption{}
         \label{fig:drums_errormap}
     \end{subfigure}
     \hfill
     \begin{subfigure}[b]{0.55\textwidth}
      \centering
       \includegraphics[width=0.9\linewidth]{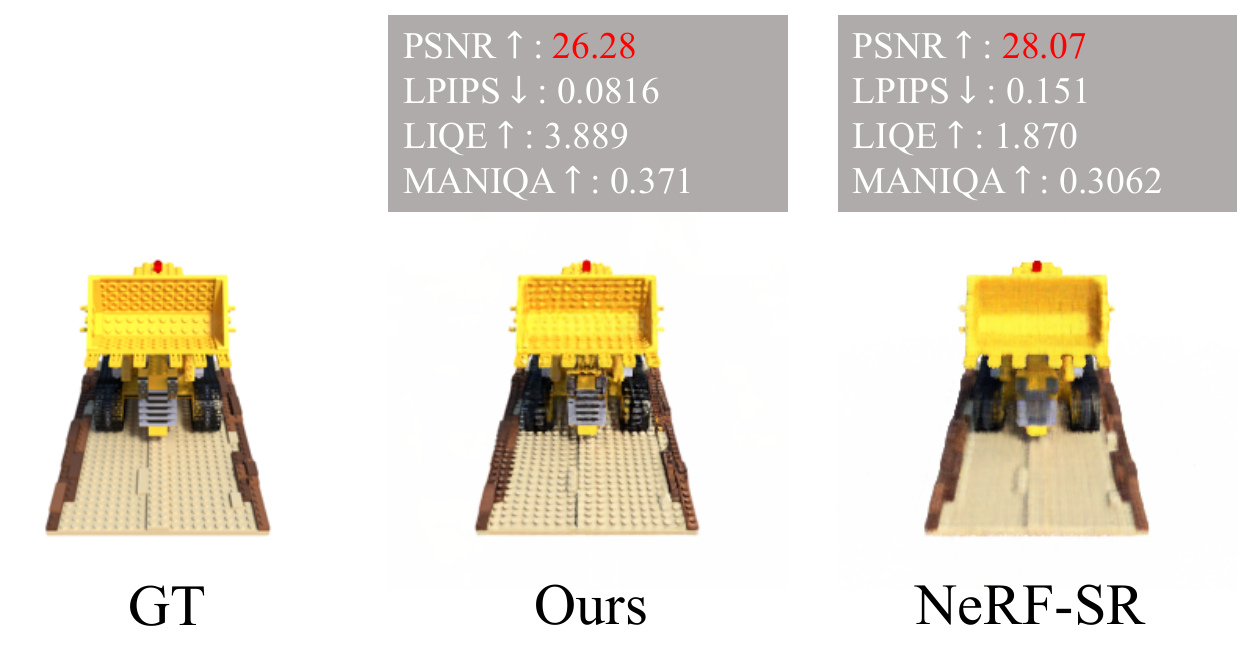}
       \caption{}
       \label{fig:psnr_and_others_compare}
     \end{subfigure}
     \vspace{-1.0em}
     \caption{(a) Error map between ground truth and our method/NeRF-Perframe. (b) we showcase that the visual quality can be much better even with lower PSNR scores.}
     \vspace{-1.0em}
\end{figure}

\noindent\textbf{Quantitative Results.}
We conduct extensive quantitative comparisons with various baselines across different restoration task in Tab. \ref{tab:super_resolution_comparison} for super-resolution,  Tab. \ref{tab:blur_consistent} and Tab. \ref{tab:blur_inconsistent} for deblurring, Tab. \ref{tab:noise} for denoising and Tab. \ref{tab:mixed_degraded_comparison} for mixed degradation. As analyzed before, although most of the time our method falls slightly behind on the reconstruction metrics like PSNR and SSIM compared with other baselines, which only measure local pixel-aligned similarity between the rendered novel views and the ground truth images, are less indicative since uncertainties naturally exist in generative procedure. Taking the super-resolution results on Blender data as an example (Tab.~\ref{tab:super_resolution_comparison}), the simplest baseline NeRF-Perframe have already achieved the best reconstruction metrics, but as shown in Figure. \ref{fig:drums_errormap}, its visual quality is vastly inferior compared with our results. Through the error map, we found the mis-alignment between the generated 3D and input 3D causes the drop of PSNR and SSIM.  By contrast, on the perceptual metrics like LPIPS and non-reference based metrics including LIPE and MANIQA, which could more effectively reflect the restoration performance, our method consistently achieves better results when compared with other baselines, demonstrating its clear advantages.

For mixed degradation tasks, the best results for LPIPS metrics are achieved by NeRFLix w. ref~\cite{zhou2023nerflix}. This is because NeRFLiX can see two high-quality ground-truth images from the nearest two viewpoints when inference, which leads to information leakage. However, high-quality ground-truth information is not accessed in our setting or any real-world cases. After eliminating the impact of ground truth (NeRFLiX w/o. ref) by replacing the reference ground truth images with degraded images, our method performs better than NeRFLiX. 

\begin{figure}
  \centering
    \includegraphics[width=\textwidth]{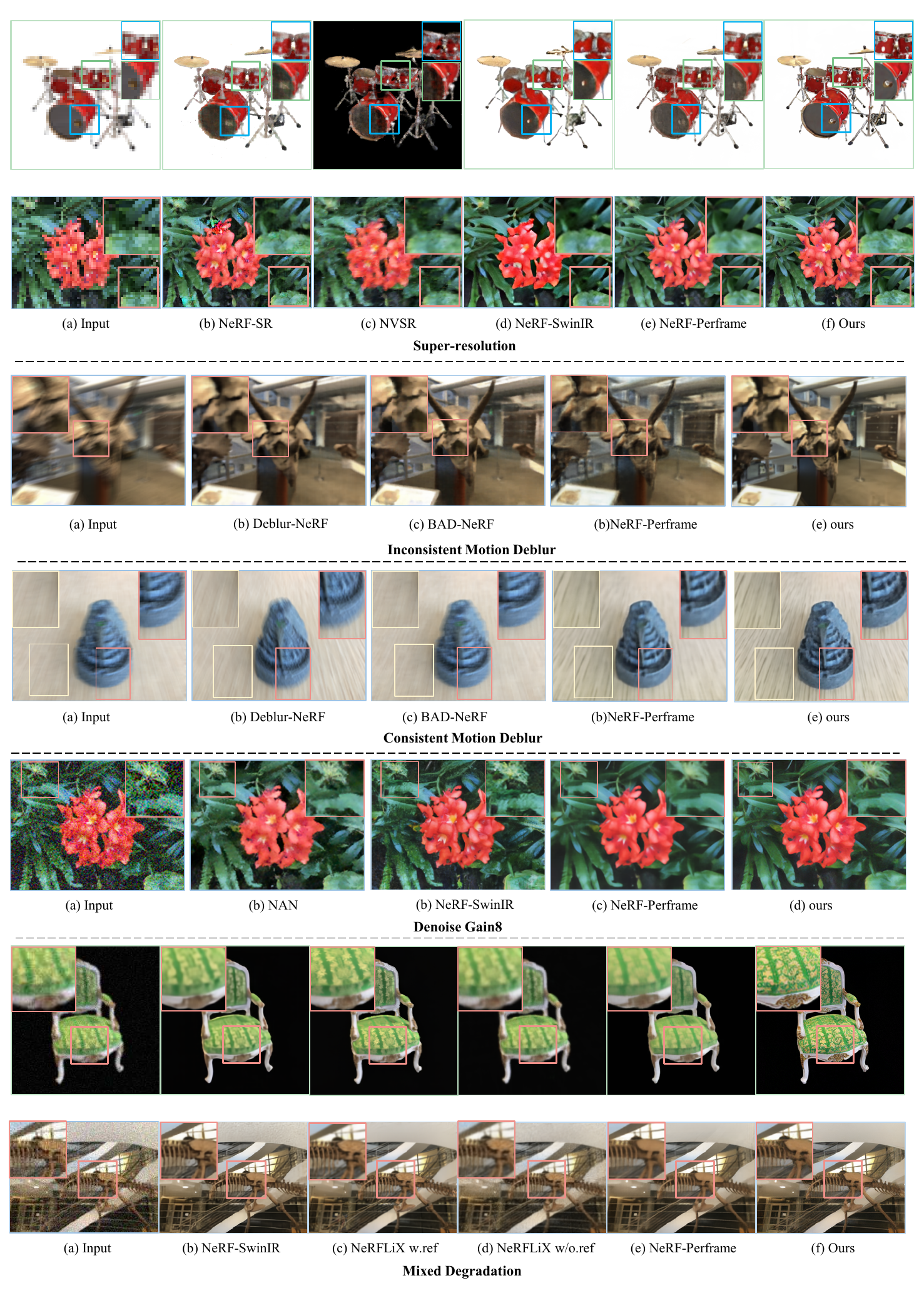}
    \caption{\textbf{Visual results for super-resolution and mixed-degradation tasks}. We show that our method is capable of generating detailed geometry and texture while other methods tend to be smooth in both geometry and texture. We recommend zooming in for better visualization.}
    \label{fig:all_visual_results}
\end{figure}

\noindent\textbf{Qualitative Results.} To further verify the restoration capability, we present visual results for different degradations in Fig. \ref{fig:all_visual_results}. Our method is able to generate realistic details while other methods tend to generate smooth results which lack high-quality details. For example, for the super-resolution task, we show the drum restoration for each method, as can be seen in  Fig. \ref{fig:all_visual_results}, our method generates high-fidelity drum surfaces with sharp edges, while other methods suffer from severe blur on the drum surfaces and edges. For mixed degradation tasks, our method successfully restores the intricate golden textures on the chair surface while other methods only rendered extremely blurred texture.

\begin{table}[!t]
\begin{subtable}[t]{0.493\textwidth}
\resizebox{\linewidth}{!}{
  \begin{tabular}{c|ccccc|ccccc}
    \toprule
     & \multicolumn{5}{|c|}{Blender} & \multicolumn{5}{|c}{LLFF}               \\
    Method & 
    PSNR$\uparrow$ &  SSIM$\uparrow$ & LPIPS$\downarrow$ & LIPE$\uparrow$ & MANIQA $\uparrow$ & 
    PSNR$\uparrow$ & SSIM$\uparrow$ & LPIPS$\downarrow$ & LIPE$\uparrow$ & MANIQA$\uparrow$\\
    \midrule
  
    NeRF-SR    & 26.85  &  0.912  & 0.135      & 2.372  & 0.366   & 22.65     & 0.702     & 0.327  & 1.320  & 0.220 \\
    
    NVSR        & 26.13             & 0.879                 & 0.132       &  2.301 & 0.343	     & 21.29     & 0.606     & 0.442   & 2.108	& 0.177	\\
    
    NeRF-SwinIR         & 24.07             & 0.878                 & 0.119    & 3.06 &	0.381      & 21.73     & 0.655     & 0.365  & 1.808 & 0.267\\

      NeRF-Perframe        & \cellcolor{pink}27.39    & \cellcolor{pink}0.922        & 0.083    &   3.276  & 0.386  & 23.68     & 0.745     & 0.261 & 1.417 & 0.257  \\
    
    \midrule
    Ours  & 24.99 & 0.901 & \cellcolor{pink}0.062 & \cellcolor{pink}4.621	 & \cellcolor{pink}0.543 & \cellcolor{pink}23.85 &  \cellcolor{pink}0.752 & \cellcolor{pink}0.197 & \cellcolor{pink}2.397 & \cellcolor{pink}0.322 \\
    \bottomrule
  \end{tabular}}
  \caption{Super-resolution}
  \label{tab:super_resolution_comparison}
\end{subtable}
\begin{subtable}[t]{0.493\textwidth}
  \resizebox{\linewidth}{!}{
  \begin{tabular}{c|ccccc|ccccc}
    \toprule
     & \multicolumn{5}{|c|}{Blender} & \multicolumn{5}{|c}{LLFF}               \\
    Method & 
    PSNR$\uparrow$ &  SSIM$\uparrow$ & LPIPS$\downarrow$ & LIPE$\uparrow$ & MANIQA $\uparrow$ & 
    PSNR$\uparrow$ & SSIM$\uparrow$ & LPIPS$\downarrow$  & LIPE$\uparrow$ & MANIQA$\uparrow$ \\
    \midrule
    NeRFLiX W. Ref  & 27.31 & 0.933 & 0.066  & 2.643	& 0.358  & 28.18 & 0.885 & 0.145  &  1.793	& 0.233 \\
    NeRFLiX W/O. Ref    & 25.78 & 0.905 & 0.107  & 1.102 & 0.166	 & \cellcolor{pink}26.28 & 0.821 & 0.290   & 1.129 & 0.183  \\ 
    NeRF-SwinIR   & \cellcolor{pink}27.42 & 0.922 & 0.086   & 2.441	& 0.317 & 25.94 & 0.813 & 0.249  & 2.013 & 0.193 \\
     NeRF-Perframe  & 26.79 & \cellcolor{pink}0.927 & 0.088 & 2.289 &0.355 & 24.45 & 0.812 & 0.267 & 1.41 & 0.257  \\
    \midrule
    Ours  & 25.28 & 0.907 & \cellcolor{pink}0.076  & \cellcolor{pink}3.947	& \cellcolor{pink}0.541 & 24.81 & \cellcolor{pink}0.832 & \cellcolor{pink}0.217 & \cellcolor{pink}2.210	& \cellcolor{pink}0.296 \\
    \bottomrule
  \end{tabular}}
  \caption{Mixed degradation}
  \label{tab:mixed_degraded_comparison}
\end{subtable}
\vspace{-0.75em}
\caption{Quantitative comparisons on super-resolution and mixed degradation tasks. The best result without using reference is highlighted. Our method achieves great superiorities on perceptual metrics and image quality when compared with others.}
\vspace{-1em}
\end{table}


\begin{table}[!t]
    \vspace{-0.5em}
    \begin{subtable}[t]{0.32\textwidth}
          \resizebox{1.0\linewidth}{!}{
              \begin{tabular}{c|ccccc}
                \toprule
                 & \multicolumn{5}{|c}{LLFF}          \\
                Method & PSNR$\uparrow$ &  SSIM$\uparrow$ & LPIPS$\downarrow$ & LIPE$\uparrow$ & MANIQA$\uparrow$ \\
                \midrule
                Deblur-NeRF & 23.75	& 0.799	& 0.307  & \cellcolor{pink}1.365 & 0.157	 \\
                BAD-NeRF & \cellcolor{pink}24.027	& 0.788	& 0.313  & 1.094 & 0.151	\\
                NeRF-Perframe &21.02 & 0.695 & 	0.362 & 0.362	& 0.150  \\
                \midrule
                Ours & 23.23 &\cellcolor{pink}0.811 & \cellcolor{pink}0.294  & 1.144 & \cellcolor{pink}0.177 \\
                \bottomrule
              \end{tabular}}
        \subcaption{Deblurring}
        \label{tab:blur_inconsistent}
    \end{subtable}
    \begin{subtable}[t]{0.32\textwidth}
        \resizebox{1.0\linewidth}{!}{
          \begin{tabular}{c|ccccc}
            \toprule
             & \multicolumn{5}{|c}{LLFF}          \\
            Method & PSNR$\uparrow$ &  SSIM$\uparrow$ & LPIPS$\downarrow$ & LIPE$\uparrow$ & MANIQA$\uparrow$ \\
            \midrule
            Deblur-NeRF & 21.71 & 0.749 & 0.333  & 1.104 & 0.122\\
            BAD-NeRF & \cellcolor{pink}25.40 & \cellcolor{pink}0.836 & 0.278  & 1.140	& 0.173 \\
            NeRF-Perframe & 21.57 & 0.771 & 0.310  & 1.103 &	0.199	 \\
            \midrule
            Ours & 22.93 & 0.789 & \cellcolor{pink}0.252  & \cellcolor{pink}1.143 & \cellcolor{pink}0.224 \\
            \bottomrule
          \end{tabular}}
        \subcaption{Deblurring (consistent blur)}
        \label{tab:blur_consistent}
    \end{subtable}
    \begin{subtable}[t]{0.32\textwidth}
          \resizebox{1.0\linewidth}{!}{
              \begin{tabular}{c|ccccc}
                \toprule
                 & \multicolumn{5}{|c}{LLFF}          \\
                Method & PSNR$\uparrow$ &  SSIM$\uparrow$ & LPIPS$\downarrow$  & LIPE$\uparrow$ & MANIQA$\uparrow$\\
                \midrule
                NAN & \cellcolor{pink}25.99 & \cellcolor{pink}0.822 & 0.3208  & 1.3032	& 0.241	 \\
                NeRF-SwinIR & 24.37 & 0.778 & 0.346 & 1.579 & 0.210 \\
                NeRF-Perframe & 23.66 &	0.786 &	0.281  & 1.095 & 0.214\\
                \midrule
                Ours & 23.78 & 0.791 & \cellcolor{pink}0.2561 & \cellcolor{pink}1.607 & \cellcolor{pink}0.257 \\
                \bottomrule
              \end{tabular}}
        \subcaption{Denoising with Gain 8}
        \label{tab:noise}
    \end{subtable}
    \vspace{-1.0em}
    \caption{Quantitative comparisons for deblurring and denoising. The best result without using reference is highlighted.  Our method performs the best on perceptual metrics.}
    \vspace{-2.0em}
\end{table}

\begin{figure}[t]
\begin{subfigure}[c]{0.42\textwidth}
  \parbox[][7.5em][c]{\linewidth}{ 
  \centering
  \includegraphics[width=\linewidth]{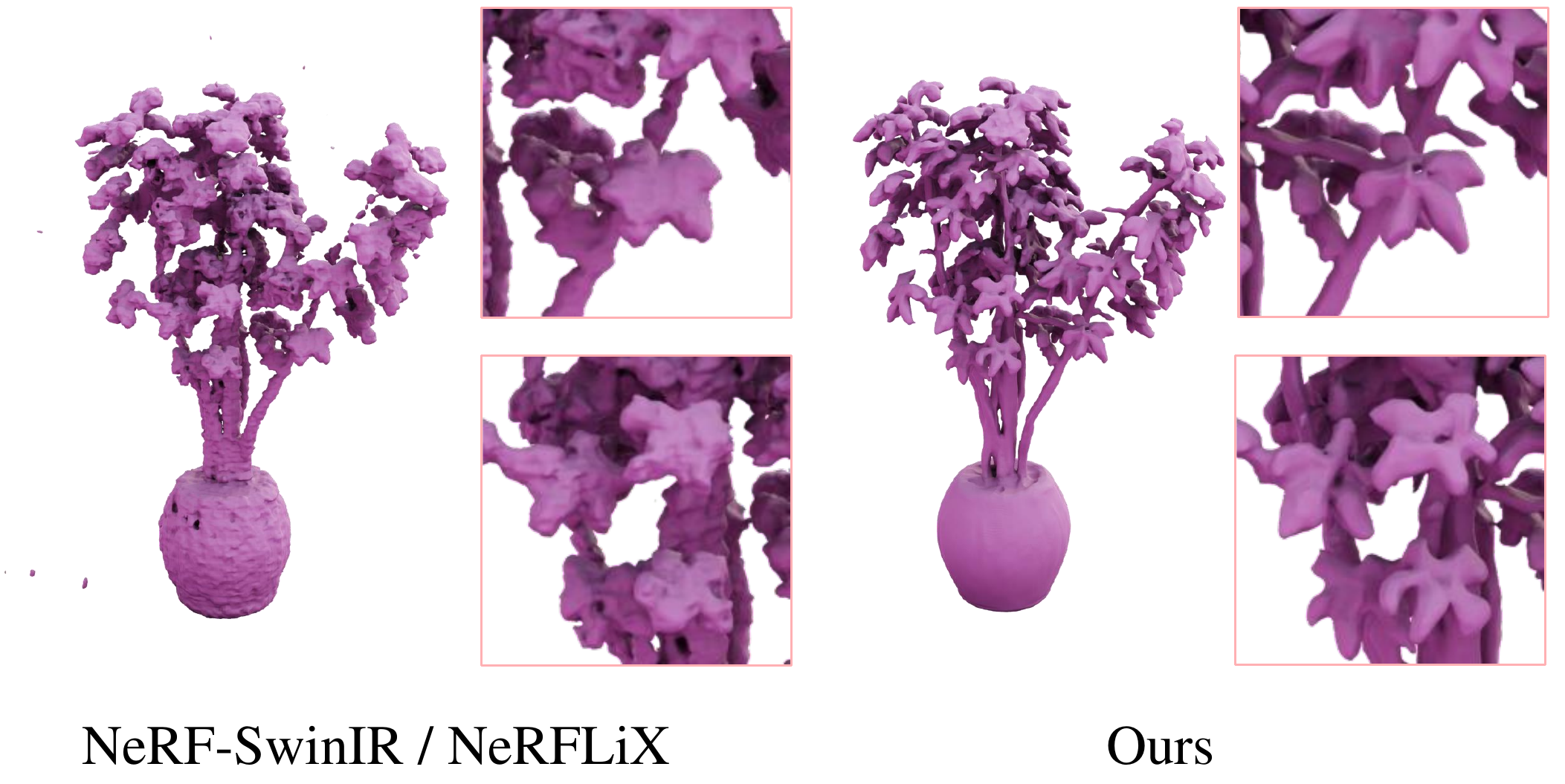}
  }
  
  \subcaption{}
  \label{fig:mesh_compare}
\end{subfigure}
\hfill
\begin{subfigure}[c]{0.51\textwidth}
  \vspace{-0.35em}
  \parbox[][7.5em][c]{\linewidth}{
  \centering
   \includegraphics[width=\linewidth]{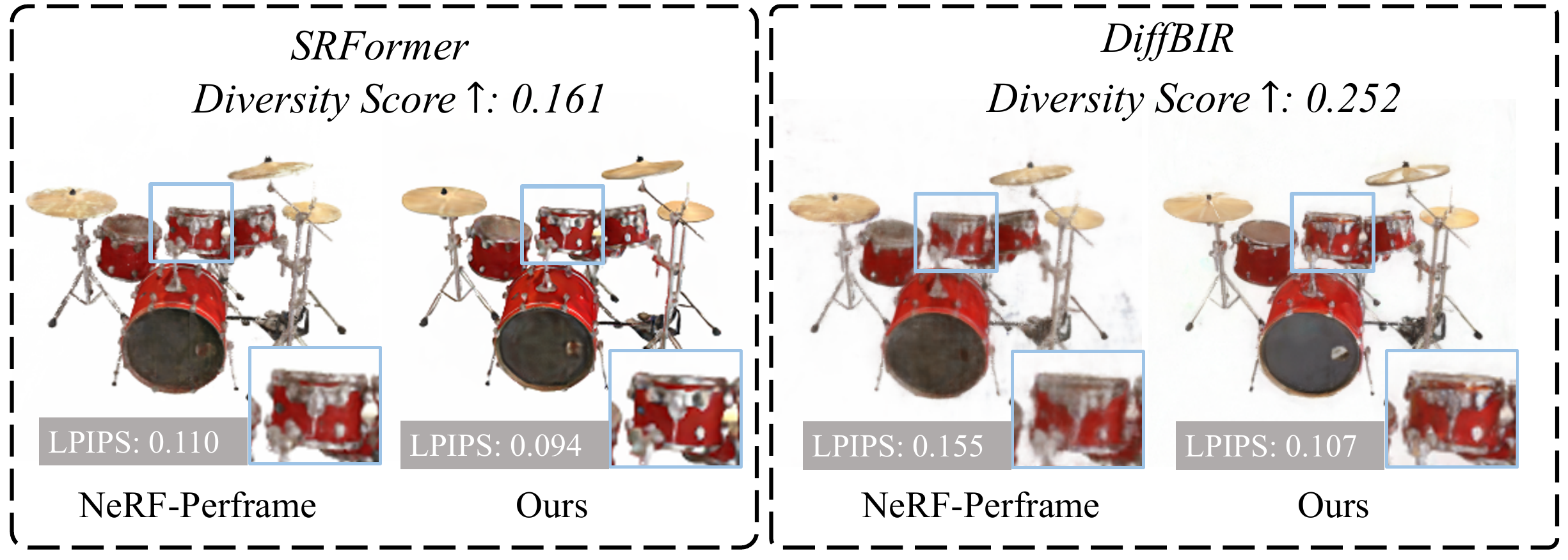}
   }
   \vspace{0.35em}
   \subcaption{}
   \label{fig:transformer_based}
\end{subfigure}
\vspace{-1em}
\caption{\textbf{(a)}Geometry comparisons between NeRF-SWINIR/NeRFLIX and our method. \textbf{(b)}Comparisons between using different 2D restoration models.}
\vspace{-1em}
\end{figure}

\noindent\textbf{Geometry Refinement.}
Compared with previous NeRF restoration methods, most of which focusing on using 2D refinement to resolve frame-wise defects, our method firstly leverages the general priors of large foundation models and GAN, achieving the open-domain 3D-based restoration. Thus, our restoration algorithm could not only recover the high-quality and view-consistent images, but also refine the 3D geometry. We demonstrate this advantage using the Ficus data of Blender, as shown in Fig. \ref{fig:mesh_compare}, through comparing the extracted mesh from the restored NeRF models, our method effectively learns the restoration in the 3D tri-plane space and recover better geometry.


\begin{figure}[t]
     \centering
     \begin{subfigure}[b]{0.47\textwidth}
        \centering
        \includegraphics[width=1.0\linewidth]{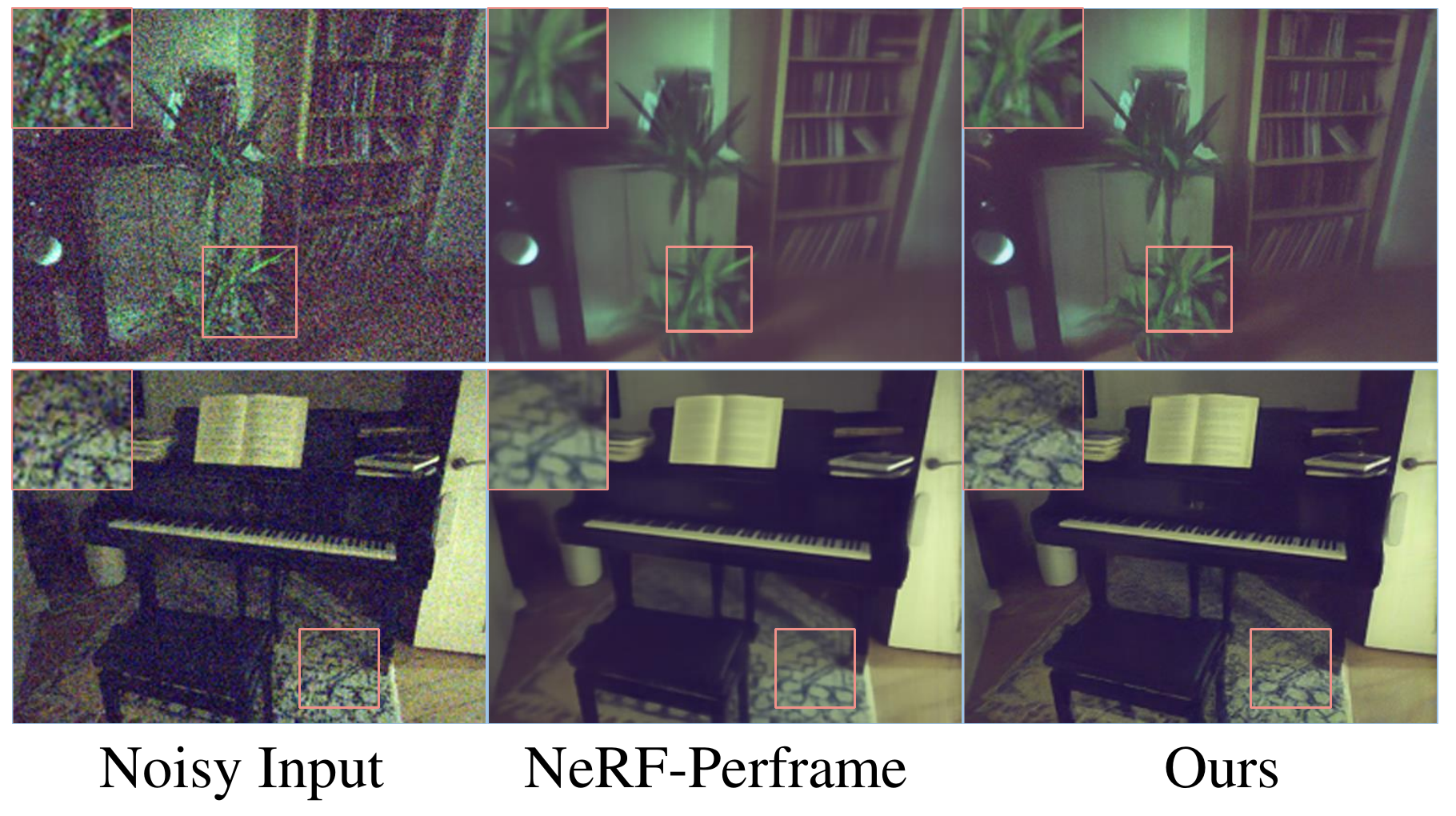}
        \caption{Noisy Input}
        \label{fig:real_noise}
     \end{subfigure}
     \hfill
     \begin{subfigure}[b]{0.52\textwidth}
        \vfill
        \centering
        \includegraphics[width=1.0\linewidth]{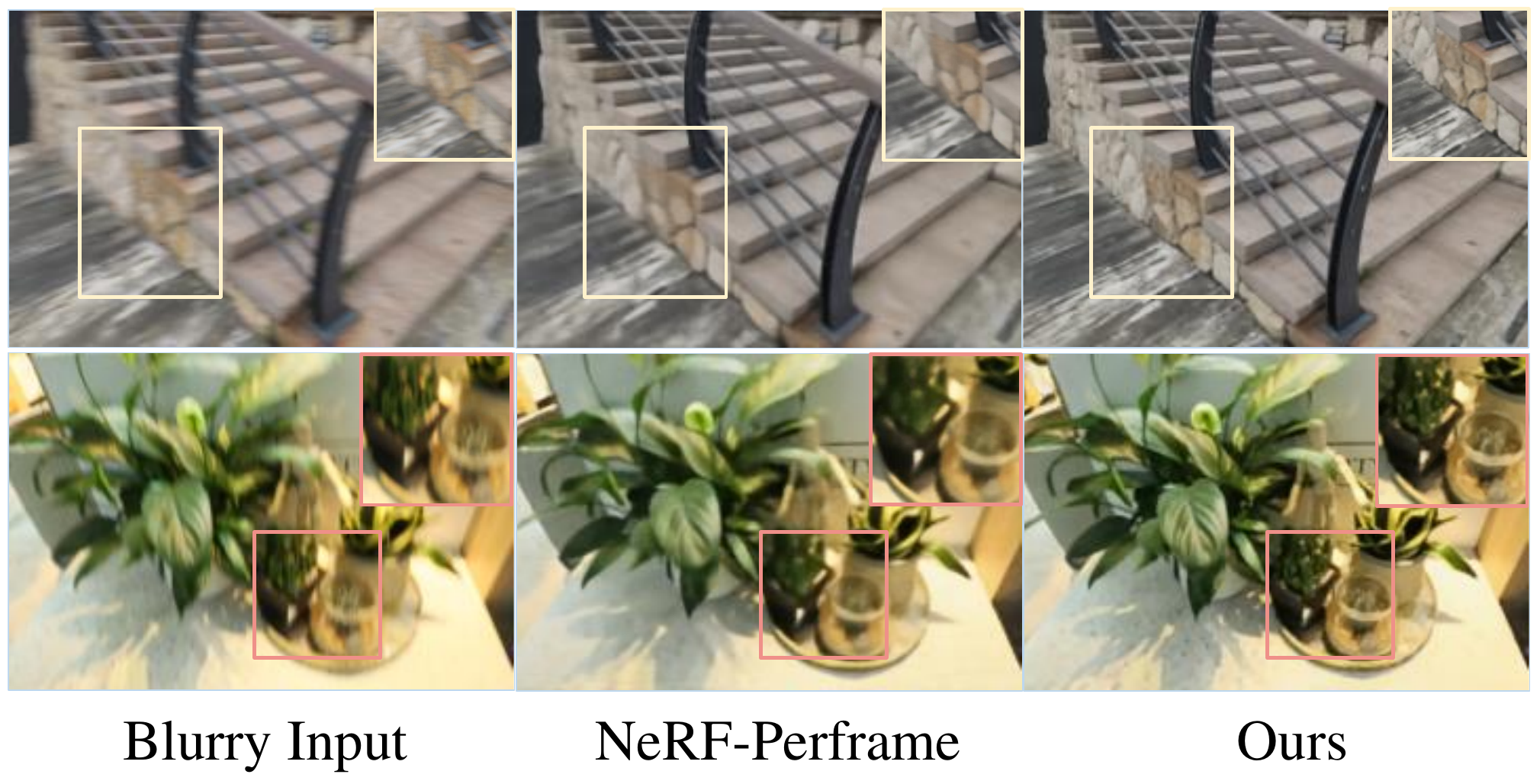}
        \caption{Blurry Input}
        \vfill
        \label{fig:realblur}
     \end{subfigure}
     \vspace{-2em}
     \caption{\textbf{The performance of our method under real world scenario.} The results indicate that our method could also generalize to the real-world setting pretty well.}
     \vspace{-1.25em}
\end{figure}

\vspace{-0.25em}
\subsection{Real-world Restoration}
To validate that our method also works well on the real-world setting, we also test RaFE using real-noise and real-blur datasets proposed in~\cite{mildenhall2022rawnerf} and~\cite{ma2022deblur}. As shown in Fig. \ref{fig:real_noise} and Fig. \ref{fig:realblur}, benefiting from the powerful 2D restoration models, NeRF-Perframe could effectively remove the existing degradations like noise or blur. Nonetheless, simply averaging the view-inconsistent 2D restored frames results in very smooth 3D reconstruction. By contrast, through modeling the 3D space using a generative model, the sampled 3D model from our method could achieve significantly better rendering quality with realistic and degradation-free texture details, demonstrating its superiorities on the real-world 3D restoration.

\vspace{-0.25em}
\subsection{Discussion}
\vspace{-0.5em}

\noindent\textbf{Effects of different restoration models.}
To investigate the influence of different 2D restoration models on our method, we tested two additional off-the-shelf restoration models for the super-resolution task, including diffusion-based DiffBIR~\cite{lin2023diffbir} and non-diffusion-based SRFormer~\cite{zhou2023srformer}. As shown in Fig. \ref{fig:transformer_based}, diffusion-based model DiffBIR shows larger restoration diversity over SRFormer as expected by measuring the diversity score. When the repaired images exhibit diversity, direct reconstruction inevitably leads to the blurriness to varying degrees due to the exisiting multi-view inconsistency. By contrast, through modeling the distribution of the potential high-quality NeRFs, our method successfully accommodates these inconsistencies and consistently achieves better performance over NeRF-Perframe, demonstrating the great generalization capability of RaFE to different 2D restoration model.

\begin{figure}[t]
    \begin{subfigure}[c]{0.74\textwidth}
    \parbox[][8.5em][c]{\linewidth}{
    \centering
    \includegraphics[width=0.95\textwidth]{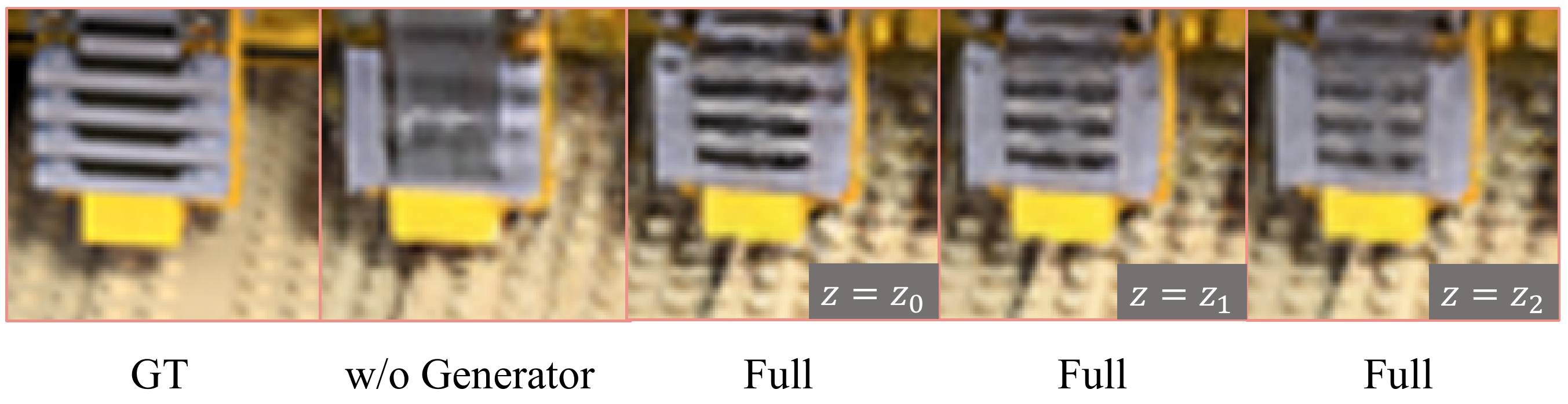}
    }
    \vspace{-1.0em}
    \end{subfigure}
    \hfill
     \begin{subtable}[c]{0.25\textwidth}
        \parbox[][8.5em][c]{\linewidth}{
        \vspace{-1em}
        \resizebox{1.0\linewidth}{!}{
            \begin{tabular}{c|cc}
            \toprule
            Generator & \color{Red}{\XSolidBrush} &   \color{ForestGreen}{\CheckmarkBold} \\
            \midrule

            LPIPS$\downarrow$ & 0.091 &  \cellcolor{pink}0.080 \\
            LIPE$\uparrow$ & 3.266 &  \cellcolor{pink}4.224 \\
            MANIQA$\uparrow$ & 0.278 &  \cellcolor{pink}0.447 \\
            \bottomrule
        \end{tabular}}
        }
        \vspace{-1.0em}
        \label{tab:coarse_nerf_ablation}
     \end{subtable}
\vspace{-1em}
\caption{\textbf{Effectiveness of the tri-plane generator}.\textbf{Left:} image rendered by NeRF without generator and images rendered by generative NeRF under different random codes $\boldsymbol{z}(\boldsymbol{z_0},\boldsymbol{z_1}, \boldsymbol{z_2})$. \textbf{Right:} numerical metrics to evaluate the efficacy of the generator. The results show the effectiveness of using the generator to model the distribution.}
\label{fig:coarse_nerf_ablation}
\vspace{-2.0em}
\end{figure}

\noindent\textbf{Effects of the generator.}
In this ablation study, we examine the influence of the generator by comparing with the baseline that directly optimizes the NeRF parameters using GAN loss and LPIPS mentioned above on the Lego dataset. As we can observe in Fig. \ref{fig:coarse_nerf_ablation}, the image rendered by generative NeRF exhibits varied fine-textured details under different random code $\boldsymbol{z}$. Once the generator is removed, the rendered images will contain blurry and smooth appearance, showing the importance of using generator to model the distribution, which can also be demonstrated by the numerical metrics in the right part of Fig. \ref{fig:coarse_nerf_ablation}.

\begin{figure}
     \vspace{-2.5em}
     \centering
     \begin{subfigure}[c]{0.68\textwidth}
        \parbox[][8.5em][c]{\linewidth}{
        \centering
        \includegraphics[width=\linewidth]{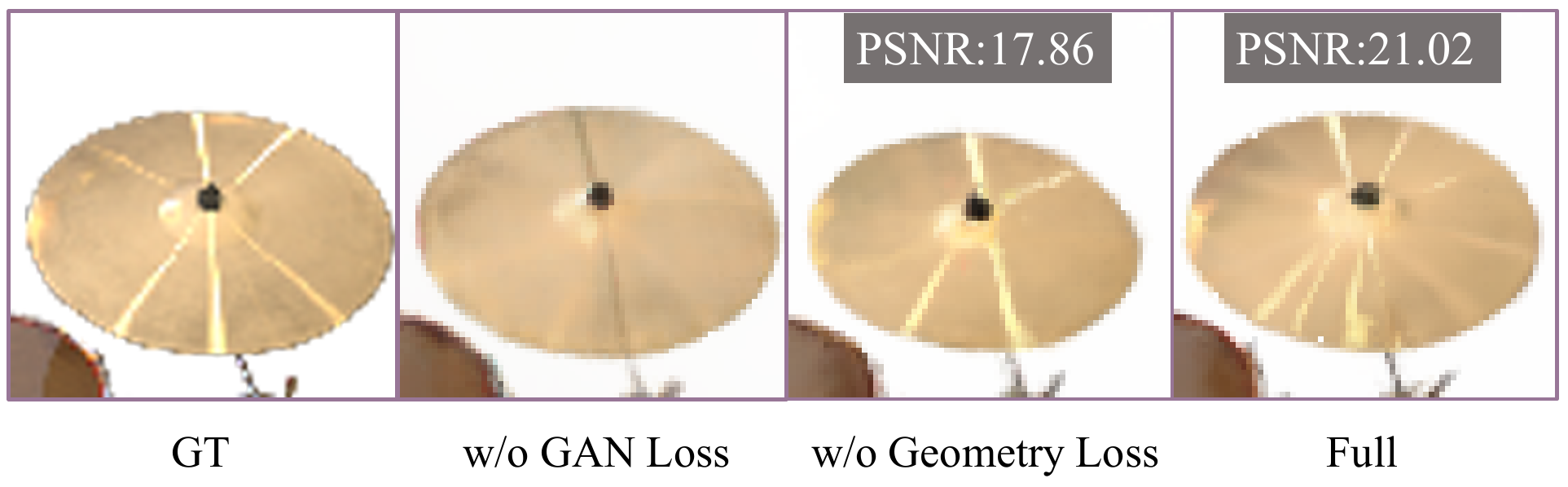}
        }
        \vspace{-0.6em}

     \end{subfigure}
     \hfill
     \begin{subtable}[c]{0.25\textwidth}
        \vspace{-0.5em}
        \parbox[][8.5em][c]{\linewidth}{
        \resizebox{1.0\linewidth}{!}{
            \begin{tabular}{c|ccc}
            \toprule
            Geometry loss & \color{ForestGreen}{\CheckmarkBold} & \color{Red}{\XSolidBrush} &   \color{ForestGreen}{\CheckmarkBold} \\
            GAN loss &\color{Red}{\XSolidBrush} & \color{ForestGreen}{\CheckmarkBold} &  \color{ForestGreen}{\CheckmarkBold} \\
            \midrule
            PSNR$\uparrow$ &  \cellcolor{pink}22.18 & 18.93 & 20.72 \\
            SSIM$\uparrow$ &  \cellcolor{pink}0.861 & 0.826 & 0.858 \\
            LPIPS$\downarrow$ & 0.190 & 0.098 &  \cellcolor{pink}0.077\\
            LIPE$\uparrow$ & 1.029 & 4.715 &  \cellcolor{pink}4.939 \\
            MANIQA$\uparrow$ & 0.157 & 0.528 &  \cellcolor{pink}0.553 \\
            \bottomrule
        \end{tabular}}
        }
        \vspace{0.1em}
        \label{tab:ablation_lpipsgan}
     \end{subtable}
     \vspace{-1em}
     \caption{\textbf{Effectiveness of geometry loss \& GAN loss term.} We show the visual results (left)  and numerical metrics (right) of using different loss terms. RaFE trained with both objectives achieves the best performance.}
     \label{fig:ablation_lpipsgan}
     \vspace{-3em}
\end{figure}

\noindent\textbf{Effects of geometry loss \& GAN loss.}
In this experiment, we examine the influence of geometry and GAN losses on the performance by training the model on the drums dataset, shown in Fig. \ref{fig:ablation_lpipsgan}. We conduct this experiment by setting the weights of unused losses to $0$. As can be seen, removing the geometry loss results in severe geometry mismatches (e.g. the edges of the drums exhibit distortion), which can be also demonstrated by the drop of PSNR. Meanwhile, using perceptual loss only makes the rendered image to be very smooth with fewer details (e.g. the light and reflection), resulting in a significant decline on perceptual metrics. RaFE trained with both objectives achieves the best performance.

\vspace{-0.25em}
\section{Conclusions}
\vspace{-1em}
This paper proposes a novel generic NeRF restoration method that applies to various types of degradations, such as low resolution, blurriness, noise, and mixed degradation. The proposed method leverages the off-the-shelf image restoration methods to restore the multi-view input images individually. To tackle the geometric and appearance inconsistencies presented in multi-view images due to individual restoration, we propose to train a GAN for NeRF generation, where a two-level tri-plane structure is adopted. The coarse-level tri-plane pre-trained by low-quality images remains fixed, while the fine-level residual tri-plane to be added to the coarse level is modeled by a GAN-based generator to capture variations in restoration. Extensive experiments on various restoration tasks with both synthetic and real cases demonstrate the superior performance of our method.

\noindent\textbf{Limitations and Future Work:}
One limitation of our method is instability when performing the restoration at extremely high resolutions, such as 4k, due to the patch-rendering strategy. Moreover, due to the inherent slow efficiency of NeRF rendering, currently the long training time also needs to be optimized. To overcome these limitations, a potential solution would be to integrate more efficient rendering techniques like Gaussian splatting\cite{kerbl3Dgaussians} into RaFE, enabling the rendering of entire images instead of using patches in real-time. We plan to resolve these issues in the future work.

\bibliographystyle{splncs04}
\bibliography{main}

\begin{thebibliography}{10}
\providecommand{\url}[1]{\texttt{#1}}
\providecommand{\urlprefix}{URL }
\providecommand{\doi}[1]{https://doi.org/#1}

\bibitem{bahat2020explorable}
Bahat, Y., Michaeli, T.: Explorable super resolution. In: Proceedings of the IEEE/CVF Conference on Computer Vision and Pattern Recognition. pp. 2716--2725 (2020)

\bibitem{bahat2022neural}
Bahat, Y., Zhang, Y., Sommerhoff, H., Kolb, A., Heide, F.: Neural volume super-resolution. arXiv preprint arXiv:2212.04666  (2022)

\bibitem{barron2021mipnerf}
Barron, J.T., Mildenhall, B., Tancik, M., Hedman, P., Martin-Brualla, R., Srinivasan, P.P.: Mip-nerf: A multiscale representation for anti-aliasing neural radiance fields (2021)

\bibitem{barron2022mipnerf360}
Barron, J.T., Mildenhall, B., Verbin, D., Srinivasan, P.P., Hedman, P.: Mip-nerf 360: Unbounded anti-aliased neural radiance fields. CVPR  (2022)

\bibitem{big_vision}
Beyer, L., Zhai, X., Kolesnikov, A.: Big vision. \url{https://github.com/google-research/big_vision} (2022)

\bibitem{chan2022efficient}
Chan, E.R., Lin, C.Z., Chan, M.A., Nagano, K., Pan, B., De~Mello, S., Gallo, O., Guibas, L.J., Tremblay, J., Khamis, S., et~al.: Efficient geometry-aware 3d generative adversarial networks. In: Proceedings of the IEEE/CVF Conference on Computer Vision and Pattern Recognition. pp. 16123--16133 (2022)

\bibitem{Chen2022ECCV}
Chen, A., Xu, Z., Geiger, A., Yu, J., Su, H.: Tensorf: Tensorial radiance fields. In: European Conference on Computer Vision (ECCV) (2022)

\bibitem{chen2023dehazenerf}
Chen, W.T., Yifan, W., Kuo, S.Y., Wetzstein, G.: Dehazenerf: Multiple image haze removal and 3d shape reconstruction using neural radiance fields. arXiv preprint arXiv:2303.11364  (2023)

\bibitem{bib:mimic3d}
Chen, X., Deng, Y., Wang, B.: Mimic3d: Thriving 3d-aware gans via 3d-to-2d imitation. In: Proceedings of the IEEE/CVF International Conference on Computer Vision (ICCV) (2023)

\bibitem{chen2023hierarchical}
Chen, Z., Zhang, Y., Ding, L., Bin, X., Gu, J., Kong, L., Yuan, X.: Hierarchical integration diffusion model for realistic image deblurring. In: NeurIPS (2023)

\bibitem{chen2023dual}
Chen, Z., Zhang, Y., Gu, J., Kong, L., Yang, X., Yu, F.: Dual aggregation transformer for image super-resolution. In: ICCV (2023)

\bibitem{fridovich2023k}
Fridovich-Keil, S., Meanti, G., Warburg, F.R., Recht, B., Kanazawa, A.: K-planes: Explicit radiance fields in space, time, and appearance. In: Proceedings of the IEEE/CVF Conference on Computer Vision and Pattern Recognition. pp. 12479--12488 (2023)

\bibitem{goodfellow2014generative}
Goodfellow, I., Pouget-Abadie, J., Mirza, M., Xu, B., Warde-Farley, D., Ozair, S., Courville, A., Bengio, Y.: Generative adversarial nets. Advances in neural information processing systems  \textbf{27} (2014)

\bibitem{han2023super}
Han, Y., Yu, T., Yu, X., Wang, Y., Dai, Q.: Super-nerf: View-consistent detail generation for nerf super-resolution. arXiv preprint arXiv:2304.13518  (2023)

\bibitem{Karras2021}
Karras, T., Aittala, M., Laine, S., H\"ark\"onen, E., Hellsten, J., Lehtinen, J., Aila, T.: Alias-free generative adversarial networks. In: Proc. NeurIPS (2021)

\bibitem{Karras2019stylegan2}
Karras, T., Laine, S., Aittala, M., Hellsten, J., Lehtinen, J., Aila, T.: Analyzing and improving the image quality of {StyleGAN}. In: Proc. CVPR (2020)

\bibitem{kawar2022denoising}
Kawar, B., Elad, M., Ermon, S., Song, J.: Denoising diffusion restoration models. In: Advances in Neural Information Processing Systems (2022)

\bibitem{kerbl3Dgaussians}
Kerbl, B., Kopanas, G., Leimk{\"u}hler, T., Drettakis, G.: 3d gaussian splatting for real-time radiance field rendering. ACM Transactions on Graphics  \textbf{42}(4) (July 2023), \url{https://repo-sam.inria.fr/fungraph/3d-gaussian-splatting/}

\bibitem{lee2023dp}
Lee, D., Lee, M., Shin, C., Lee, S.: Dp-nerf: Deblurred neural radiance field with physical scene priors. In: Proceedings of the IEEE/CVF Conference on Computer Vision and Pattern Recognition. pp. 12386--12396 (2023)

\bibitem{lee2023exblurf}
Lee, D., Oh, J., Rim, J., Cho, S., Lee, K.M.: Exblurf: Efficient radiance fields for extreme motion blurred images. In: Proceedings of the IEEE/CVF International Conference on Computer Vision. pp. 17639--17648 (2023)

\bibitem{li2023real}
Li, H., Zhang, Z., Jiang, T., Luo, P., Feng, H., Xu, Z.: Real-world deep local motion deblurring. In: Proceedings of the AAAI Conference on Artificial Intelligence. vol.~37, pp. 1314--1322 (2023)

\bibitem{li2022blip}
Li, J., Li, D., Xiong, C., Hoi, S.: Blip: Bootstrapping language-image pre-training for unified vision-language understanding and generation. In: ICML (2022)

\bibitem{Li_2023_CVPR}
Li, J., Zhang, Z., Liu, X., Feng, C., Wang, X., Lei, L., Zuo, W.: Spatially adaptive self-supervised learning for real-world image denoising. In: Proceedings of the IEEE/CVF Conference on Computer Vision and Pattern Recognition (CVPR). pp. 9914--9924 (June 2023)

\bibitem{liang2021swinir}
Liang, J., Cao, J., Sun, G., Zhang, K., Van~Gool, L., Timofte, R.: Swinir: Image restoration using swin transformer. arXiv preprint arXiv:2108.10257  (2021)

\bibitem{lin2023diffbir}
Lin, X., He, J., Chen, Z., Lyu, Z., Fei, B., Dai, B., Ouyang, W., Qiao, Y., Dong, C.: Diffbir: Towards blind image restoration with generative diffusion prior. arXiv preprint arXiv:2308.15070  (2023)

\bibitem{ma2022deblur}
Ma, L., Li, X., Liao, J., Zhang, Q., Wang, X., Wang, J., Sander, P.V.: Deblur-nerf: Neural radiance fields from blurry images. In: Proceedings of the IEEE/CVF Conference on Computer Vision and Pattern Recognition. pp. 12861--12870 (2022)

\bibitem{mildenhall2018burst}
Mildenhall, B., Barron, J.T., Chen, J., Sharlet, D., Ng, R., Carroll, R.: Burst denoising with kernel prediction networks. In: Proceedings of the IEEE conference on computer vision and pattern recognition. pp. 2502--2510 (2018)

\bibitem{mildenhall2022rawnerf}
Mildenhall, B., Hedman, P., Martin-Brualla, R., Srinivasan, P.P., Barron, J.T.: {NeRF} in the dark: High dynamic range view synthesis from noisy raw images. CVPR  (2022)

\bibitem{mildenhall2019llff}
Mildenhall, B., Srinivasan, P.P., Ortiz-Cayon, R., Kalantari, N.K., Ramamoorthi, R., Ng, R., Kar, A.: Local light field fusion: Practical view synthesis with prescriptive sampling guidelines. ACM Transactions on Graphics (TOG)  (2019)

\bibitem{mildenhall2021nerf}
Mildenhall, B., Srinivasan, P.P., Tancik, M., Barron, J.T., Ramamoorthi, R., Ng, R.: Nerf: Representing scenes as neural radiance fields for view synthesis. Communications of the ACM  \textbf{65}(1),  99--106 (2021)

\bibitem{mueller2022instant}
M\"uller, T., Evans, A., Schied, C., Keller, A.: Instant neural graphics primitives with a multiresolution hash encoding. ACM Trans. Graph.  \textbf{41}(4),  102:1--102:15 (Jul 2022). \doi{10.1145/3528223.3530127}, \url{https://doi.org/10.1145/3528223.3530127}

\bibitem{pearl2022noiseaware}
Pearl, N., Treibitz, T., Korman, S.: Nan: Noise-aware nerfs for burst-denoising. In: CVPR (2022)

\bibitem{saharia2022palette}
Saharia, C., Chan, W., Chang, H., Lee, C., Ho, J., Salimans, T., Fleet, D., Norouzi, M.: Palette: Image-to-image diffusion models. In: ACM SIGGRAPH 2022 Conference Proceedings. pp. 1--10 (2022)

\bibitem{saharia2022photorealistic}
Saharia, C., Chan, W., Saxena, S., Li, L., Whang, J., Denton, E.L., Ghasemipour, K., Gontijo~Lopes, R., Karagol~Ayan, B., Salimans, T., et~al.: Photorealistic text-to-image diffusion models with deep language understanding. Advances in Neural Information Processing Systems  \textbf{35},  36479--36494 (2022)

\bibitem{simonyan2014very}
Simonyan, K., Zisserman, A.: Very deep convolutional networks for large-scale image recognition. arXiv preprint arXiv:1409.1556  (2014)

\bibitem{skorokhodov2022epigraf}
Skorokhodov, I., Tulyakov, S., Wang, Y., Wonka, P.: Epigraf: Rethinking training of 3d gans. Advances in Neural Information Processing Systems  \textbf{35},  24487--24501 (2022)

\bibitem{SunSC22}
Sun, C., Sun, M., Chen, H.: Direct voxel grid optimization: Super-fast convergence for radiance fields reconstruction. In: CVPR (2022)

\bibitem{tian2024visual}
Tian, K., Jiang, Y., Yuan, Z., Bingyue, P., Wang, L.: Visual autoregressive modeling: Scalable image generation via next-scale prediction. arXiv preprint arXiv:2404.02905  (2024)

\bibitem{wan2023cad}
Wan, Z., Paschalidou, D., Huang, I., Liu, H., Shen, B., Xiang, X., Liao, J., Guibas, L.: Cad: Photorealistic 3d generation via adversarial distillation. arXiv preprint arXiv:2312.06663  (2023)

\bibitem{Wan_2023_CVPR}
Wan, Z., Richardt, C., Bo\v{z}i\v{c}, A., Li, C., Rengarajan, V., Nam, S., Xiang, X., Li, T., Zhu, B., Ranjan, R., Liao, J.: Learning neural duplex radiance fields for real-time view synthesis. In: Proceedings of the IEEE/CVF Conference on Computer Vision and Pattern Recognition (CVPR). pp. 8307--8316 (June 2023)

\bibitem{wan2020bringing}
Wan, Z., Zhang, B., Chen, D., Zhang, P., Chen, D., Liao, J., Wen, F.: Bringing old photos back to life. In: proceedings of the IEEE/CVF conference on computer vision and pattern recognition. pp. 2747--2757 (2020)

\bibitem{wang2022nerf}
Wang, C., Wu, X., Guo, Y.C., Zhang, S.H., Tai, Y.W., Hu, S.M.: Nerf-sr: High quality neural radiance fields using supersampling. In: Proceedings of the 30th ACM International Conference on Multimedia. pp. 6445--6454 (2022)

\bibitem{wang2023badnerf}
Wang, P., Zhao, L., Ma, R., Liu, P.: {BAD-NeRF: Bundle Adjusted Deblur Neural Radiance Fields}. In: Proceedings of the IEEE/CVF Conference on Computer Vision and Pattern Recognition (CVPR). pp. 4170--4179 (June 2023)

\bibitem{wang2023bad}
Wang, P., Zhao, L., Ma, R., Liu, P.: Bad-nerf: Bundle adjusted deblur neural radiance fields. In: Proceedings of the IEEE/CVF Conference on Computer Vision and Pattern Recognition. pp. 4170--4179 (2023)

\bibitem{wang2022zero}
Wang, Y., Yu, J., Zhang, J.: Zero-shot image restoration using denoising diffusion null-space model. The Eleventh International Conference on Learning Representations  (2023)

\bibitem{wang2004image}
Wang, Z., Bovik, A.C., Sheikh, H.R., Simoncelli, E.P.: Image quality assessment: from error visibility to structural similarity. IEEE transactions on image processing  \textbf{13}(4),  600--612 (2004)

\bibitem{yang2022maniqa}
Yang, S., Wu, T., Shi, S., Lao, S., Gong, Y., Cao, M., Wang, J., Yang, Y.: Maniqa: Multi-dimension attention network for no-reference image quality assessment. In: Proceedings of the IEEE/CVF Conference on Computer Vision and Pattern Recognition. pp. 1191--1200 (2022)

\bibitem{zhang2021designing}
Zhang, K., Liang, J., Van~Gool, L., Timofte, R.: Designing a practical degradation model for deep blind image super-resolution. In: IEEE International Conference on Computer Vision. pp. 4791--4800 (2021)

\bibitem{zhang2018perceptual}
Zhang, R., Isola, P., Efros, A.A., Shechtman, E., Wang, O.: The unreasonable effectiveness of deep features as a perceptual metric. In: CVPR (2018)

\bibitem{zhang2023liqe}
Zhang, W., Zhai, G., Wei, Y., Yang, X., Ma, K.: Blind image quality assessment via vision-language correspondence: A multitask learning perspective. In: IEEE Conference on Computer Vision and Pattern Recognition. pp. 14071--14081 (2023)

\bibitem{zhang2023seal}
Zhang, W., Li, X., Chen, X., Qiao, Y., Wu, X.M., Dong, C.: Seal: A framework for systematic evaluation of real-world super-resolution. arXiv preprint arXiv:2309.03020  (2023)

\bibitem{zhou2023nerflix}
Zhou, K., Li, W., Wang, Y., Hu, T., Jiang, N., Han, X., Lu, J.: Nerflix: High-quality neural view synthesis by learning a degradation-driven inter-viewpoint mixer. In: Proceedings of the IEEE/CVF Conference on Computer Vision and Pattern Recognition. pp. 12363--12374 (2023)

\bibitem{zhou2023srformer}
Zhou, Y., Li, Z., Guo, C.L., Bai, S., Cheng, M.M., Hou, Q.: Srformer: Permuted self-attention for single image super-resolution. arXiv preprint arXiv:2303.09735  (2023)

\end{thebibliography}

\end{document}


\title{RaFE: Generative Radiance Fields Restoration \\
\textbf{\textit{Supplementary Material}}} 


\author{Zhongkai Wu\inst{1} \and
Ziyu Wan\inst{2} \and
Jing Zhang\inst{1} \and
Jing Liao\inst{2} \and
Dong Xu\inst{3}}

\authorrunning{Wu et al.}

\institute{College of Software, Beihang University, China \and
City University of Hong Kong, China \and
The University of Hong Kong, China\\
\email{ZhongkaiWu@buaa.edu.cn} \\
\href{https://zkaiwu.github.io/RaFE-Project/}{RaFE.github.io}
}

\maketitle

\section{Overview}
In this supplementary material, additional training details and more experimental results are provided, including:

\begin{itemize}
    \item Training \& rendering details of the whole pipeline in Sec. \ref{sec:training_and_rendering}.
    \item Training details of the pre-trained coarse NeRF in Sec. \ref{sec:coarse_nerf_details}.
    \item More Discussion and experimental results in Sec. \ref{sec:more_discussion}.
    \item Details of the calculation of diversity score in Sec. \ref{sec:details_diversity_score}.
    \item More real-world results in Sec. \ref{sec:more_results}.
\end{itemize}

\section{Training \& Rendering Details}
\label{sec:training_and_rendering}
The input images and corresponding viewpoints are randomly selected during the training process. We also randomly sample random code $z$ from the normal distribution. And we use a discriminator learning rate of 0.002 and a generator learning rate of 0.0025. In the early stage of training, we blur the image to stabilize the training process and reduce the blur kernel size to zero gradually. Furthermore, we use a density regularization which minimizes the density differences between adjacent sampled points.

For the NeRF-Synthetic benchmark dataset (Blender), we sample 192  points for each ray, with 128 stratified sampling points and 48 importance sampling points. We assume that the blender object is in a $[-1.5, 1.5]^3$ cube and set the near and far plane of the ray to 2 and 6 respectively. For training, we restore 10K high-quality images randomly selected in 200 training views and we set the batch size to 32 and the minibatch std group size to 4 for optimizing the generator. 

For the forward-facing datasets, we sample 192 points for each ray, with 128 stratified sampling points and 48 importance sampling points. We use normalized
device coordinates (NDC) and set the near plane to 0 and the far plane to 1. The derivation of NDC can be found in NeRF~\cite{mildenhall2021nerf}. The size of the dataset, the batch size, and the minibatch standard deviation group size are the same as those in the Blender dataset.

\section{Details of Pre-trained Coarse NeRF}
\label{sec:coarse_nerf_details}
As mentioned in the main submission, we pre-train a coarse NeRF as the foundation of our residual fine NeRF. In this section, we describe our training details about coarse NeRF. We use NeRFStudio~\cite{nerfstudio} as our codebase. We re-implement the hybrid explicit-implicit tri-plane representation within this framework and use degraded images for training supervision. We minimize the L2 loss, denoted as $L_{rec}$, between the rendered image and the degraded image. Inspired by K-planes~\cite{kplanes_2023}, we incorporate additional losses: the TV loss ($L_{tv}$), which minimizes feature differences between adjacent coordinates on the tri-planes, and the distortion loss ($L_{dis}$), which pushes the density of a ray within a specific range. The total loss for optimizing the coarse NeRF is:
\begin{equation}
    \mathcal{L}_{coarse} = \lambda_{rec}L_{rec} + \lambda_{tv}L_{tv} + \lambda_{dis}L_{dis},
  \label{eq:mimic_loss}
\end{equation}
where $\lambda_{rec}, \lambda_{tv}$ and $\lambda_{dis}$ denote the trade-off parameters. Specifically, we use $\lambda_{rec} = 1, \lambda_{tv}=0.01$ and $ \lambda_{dis}=0.001$ in our experiments.


\section{More Discussions}
\label{sec:more_discussion}

\begin{figure}[!ht]
  \centering
   \includegraphics[width=0.55\linewidth]{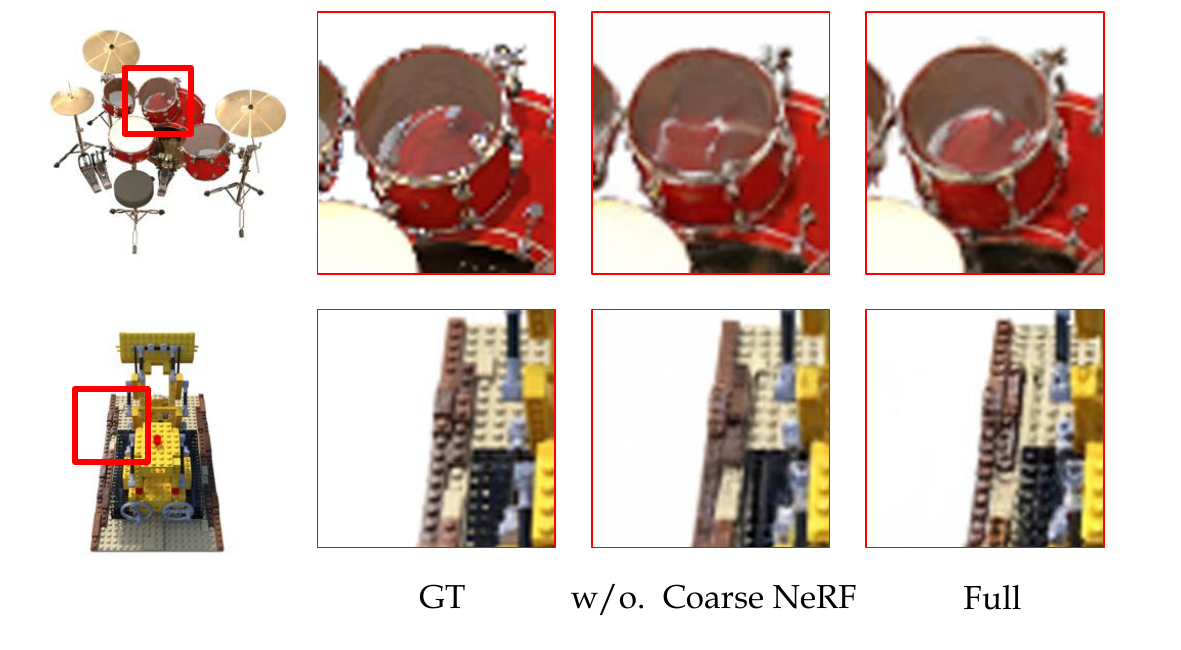}
   \caption{\textbf{Ablation study of residual coarse NeRF.} Our full pipeline renders the images with better quality, demonstrating the great effectiveness of the residual coarse NeRF.}
   \label{fig:ablation_residual}
\end{figure}

\noindent\textbf{Effects of residual coarse NeRF.}
In this ablation study, we show that the residual coarse NeRF could facilitate the model to be aware of the geometry and help better render highly detailed images, as shown in Fig. \ref{fig:ablation_residual}. We directly drop the coarse NeRF and train the generator from scratch with GAN loss and LPIPS loss, forcing the generator to learn the coarse structure and details jointly. We can observe that with the addition of coarse NeRF, RaFE could better model the structural information (like the Lego baseplate) and transparent material (like the drumhead).

\begin{figure}[!ht]
  \centering
   \includegraphics[width=0.6\linewidth]{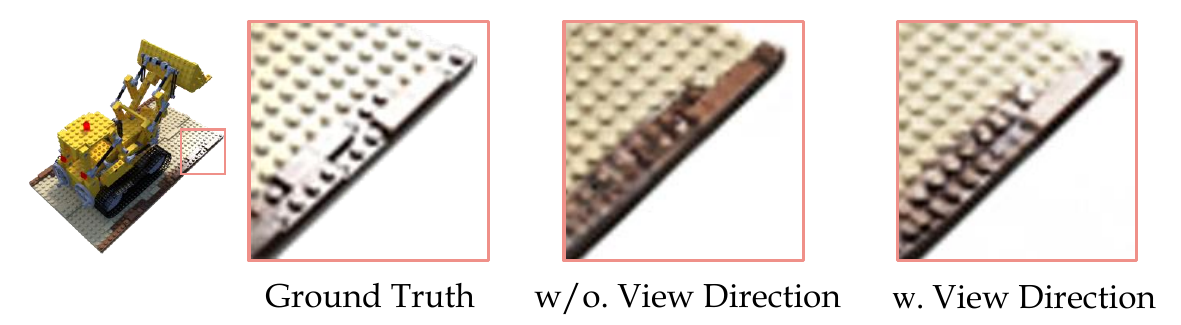}
   \caption{\textbf{Ablation of view direction.} Reflections can be observed when using view direction conditioning.}
   \label{fig:ablation_vd}
\end{figure}

\noindent\textbf{Effects of view direction.}
In this ablation study, we examine the effects of view direction conditioning. We drop the second MLP $\mathcal{M}_{color}$ in the decoder and let the RGB value directly be decoded by the first MLP jointly with density. As shown in Fig. \ref{fig:ablation_vd}, objects with non-Lambertian surfaces exhibit significant light reflections. In contrast, without view direction information, the object in the figure appears to have no reflections.

\begin{figure}[!ht]
  \centering
   \includegraphics[width=\linewidth]{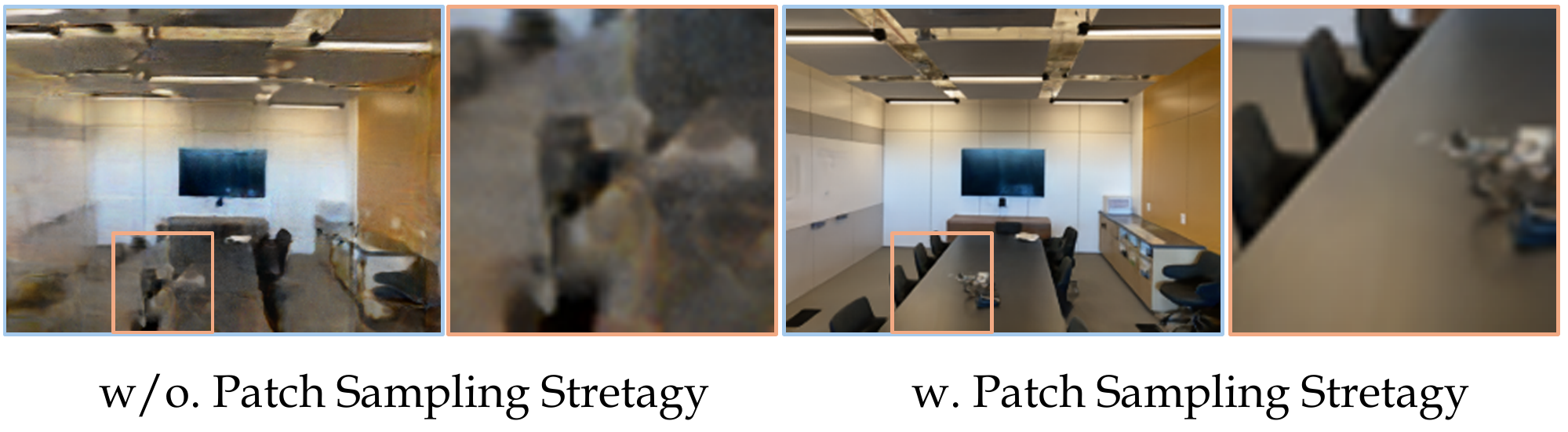}
   \caption{\textbf{Ablation of patch sampling strategy.} The training process becomes unstable without patch sampling strategy and causes severe artifacts}
   \label{fig:ablation_patch_samping}
\end{figure}

\noindent\textbf{Effects of patch sampling strategy.}
In this experiment, we examine the effects of our patch-sampling strategy by training the model on the room scene. To highlight the significance of the Beta sampling strategy, we consider the case where the value of $\delta_x, \delta_y$ are sampled from a uniform distribution: $\delta_x, \delta_y \sim \mathcal{U}(0, 1)$. This approach essentially obtains image patches through uniform cropping, which offers a baseline for comparison. The results are shown in Fig. \ref{fig:ablation_patch_samping}. As can be seen, the model trained with uniform sampling suffers from unstable optimization and severe artifacts, which effectively demonstrates the efficacy of beta sampling in maintaining the rendering quality.

\begin{figure}[!ht]
  \centering
   \includegraphics[width=0.8\linewidth]{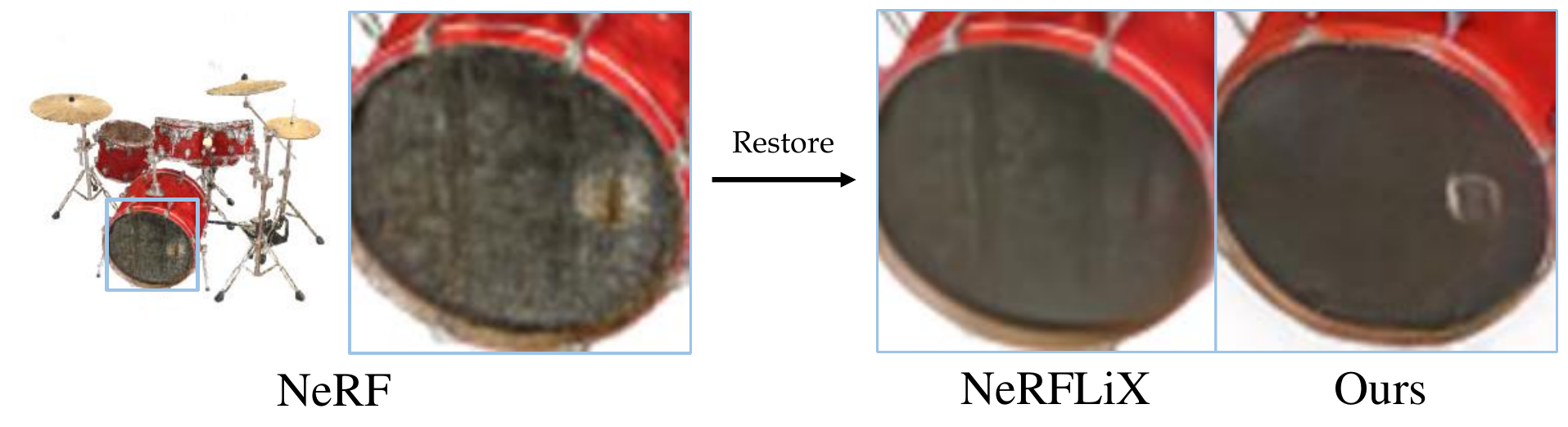}
   \caption{\textbf{Performance on NeRF-like degradation.} Although not specifically tailored for the NeRF-like degradation, our method still demonstrates satisfactory performance.}
   \label{fig:nerf_like_degradation}
\end{figure}

\noindent\textbf{Performance on NeRF-like degradation.}
In this experiment, we additionally examine our method's performance on NeRF-like degradation~\cite{zhou2023nerflix} caused by the reconstruction process of NeRF. We first use high-quality images to train a NeRF.  Although trained sufficiently, artifacts can still be presented in the images rendered by NeRF models, as discussed in ~\cite{zhou2023nerflix}. Then we collect the rendered images as a low-quality set and use our restoration method to recover an artifact-free NeRF. As can be seen in Fig. \ref{fig:nerf_like_degradation}, Although not specifically tailored for the NeRF-like degradation, our method still demonstrates clearly improved performance.

\section{Details of the diversity score}
\label{sec:details_diversity_score}
In this section, we describe the details of the diversity score used to evaluate the diversity of the generated image set. In particular, we follow the computational methods described in ~\cite{zheng2019pluralistic}. We first calculate the minimal LPIPS score for each image with other images in the image set. Then we average the per-image scores to get the overall score for the whole image set. The higher diversity score means the greater diversity of the image set. Here we provide pseudo-code in Algorithm \ref{algo:diversity_score} to explain our computational method better.  
\begin{algorithm}
    \caption{Calculation process of the diversity score}
    \label{algo:diversity_score}
    \KwIn{Image set $I$}
    \KwOut{diversity score $s$}
    \BlankLine
    \BlankLine

    $s_{sum} \leftarrow 0$ \\
    \ForEach{$i_s$ in $I$}{
        $s_{min} \leftarrow \inf$ \\
        \ForEach{$i_d$ in $I$}{
            \If{$i_s \neq i_d$}{
                $s_{lpips} \leftarrow LPIPS(i_s, i_d)$ \\
                $s_{min} \leftarrow min(s_{min}, s_{lpips})$
            }
        }
        $s_{sum} \leftarrow s_{sum} + s_{min}$
    }
    $s \leftarrow s_{sum} / \lvert A \rvert$ \\
    \KwRet{s}
    
\end{algorithm}

\section{More results of real-world scenario}
\label{sec:more_results}
\begin{figure}[!ht]
  \centering
   \includegraphics[width=\linewidth]{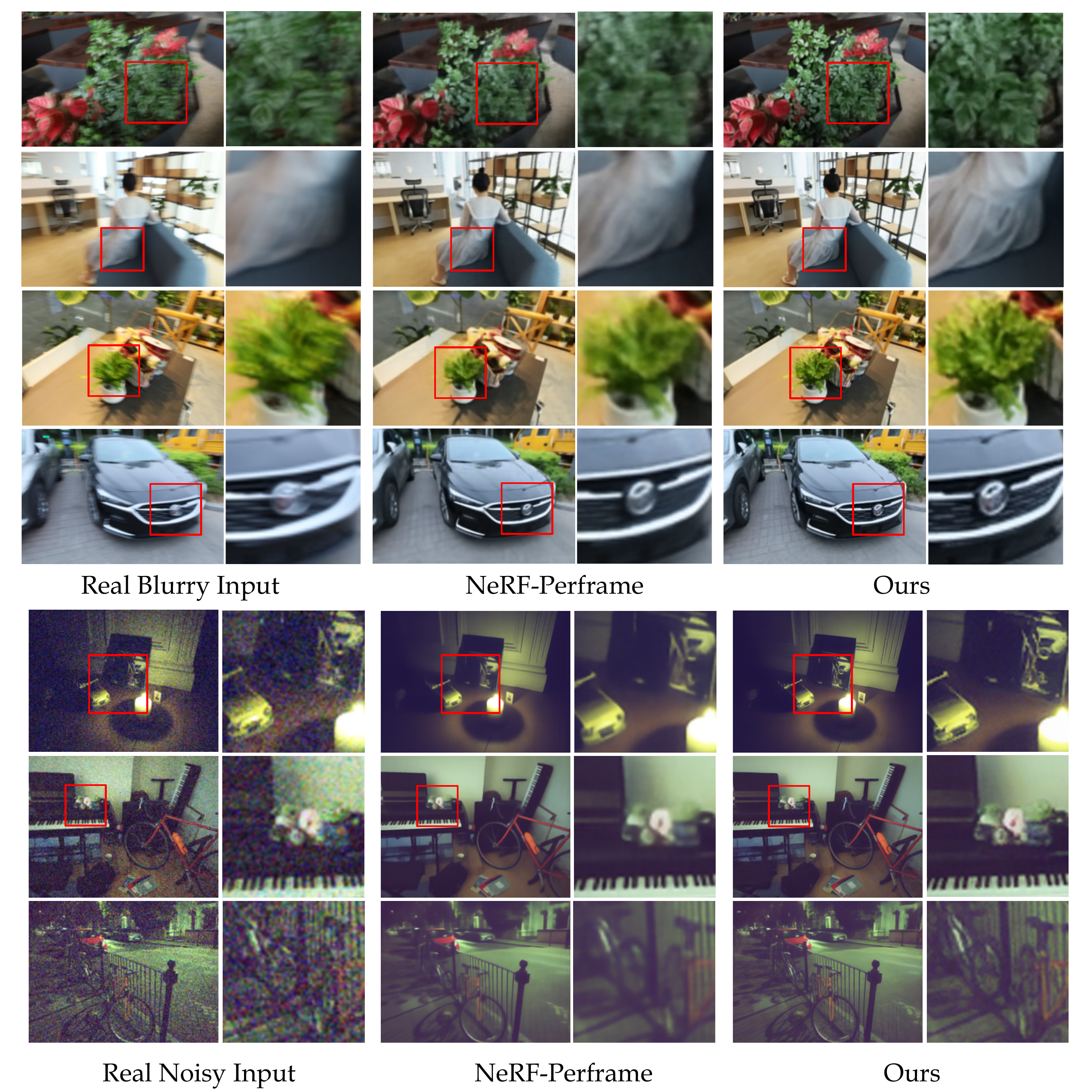}
   \caption{\textbf{More qualitative results on different real-world 3D restoration tasks. } }
   \label{fig:real_more}
\end{figure}

In this section, we provide more experiments on real-world scenarios to show the capability of our method in addressing multiple types of degradation in the real world in Fig. \ref{fig:real_more}.

\bibliographystyle{splncs04}
\bibliography{main}